\documentclass{article}

\PassOptionsToPackage{numbers, compress}{natbib}
\usepackage[eandd,preprint]{neurips_2025}
\usepackage[utf8]{inputenc}
\usepackage[T1]{fontenc}
\usepackage{hyperref}
\usepackage{url}
\usepackage{microtype}
\usepackage{xcolor}
\usepackage{longtable}

\usepackage{tikz}
\usepackage{amsmath,amssymb,amsfonts}
\usepackage{algorithmic}
\usepackage{algorithm}
\usepackage{graphicx}
\usepackage{booktabs}
\usepackage{multirow}
\usepackage{array}
\usepackage{subcaption}
\usepackage{pgfplots}
\usepackage{float}
\usepackage{wrapfig}
\usetikzlibrary{shapes,arrows,positioning,fit,calc,patterns,decorations.pathreplacing,shadows}
\pgfplotsset{compat=1.17}

\title{Manifold of Failure: Behavioral Attraction Basins in Language Models}
\title{Manifold of Failure: Behavioral Attraction Basins in Language Models\thanks{This work was conducted independently and does not reflect the views, policies, or endorsements of the
  authors' respective employers. Corresponding author: manish.bhatt13212@gmail.com}}

\author{%
  Sarthak Munshi\footnotemark[2] \\
  Amazon Web Services \\
  \And
  Manish Bhatt\footnotemark[2] \\
  Amazon Leo \\
  \And
  Vineeth Sai Narajala\footnotemark[2] \\
  Cisco \\
  \And
  Idan Habler\footnotemark[2] \\
  Cisco \\
  \And
  Ammar Al-Kahfah\footnotemark[2] \\
  Amazon Web Services \\
  \And
  Ken Huang \\
  Distributedapps.ai \\
  \And
  Blake Gatto \\
  Shrewd Security \\
}

\begin{document}

\maketitle
 \renewcommand{\thefootnote}{\fnsymbol{footnote}}
 \footnotetext[2]{Equal contribution}

 \renewcommand{\thefootnote}{\fnsymbol{footnote}}
 \footnotetext[3]{Interactive simulation available at: \href{https://manifoldoffailure.wiki/}{https://manifoldoffailure.wiki/}}

\begin{abstract}
While prior work has focused on projecting adversarial examples back onto the manifold of natural data to restore safety, we argue that comprehensive AI safety requires characterizing the unsafe regions themselves. This paper introduces a framework for systematically mapping the \emph{Manifold of Failure} in Large Language Models, reframing vulnerability search as a quality-diversity problem: MAP-Elites illuminates the continuous topology of \emph{behavioral attraction basins}, guided by an \emph{Alignment Deviation} metric that measures divergence from intended alignment. Across Llama-3-8B, GPT-OSS-20B, and GPT-5-Mini, we achieve up to $63\%$ behavioral coverage, discover up to $370$ distinct vulnerability niches, and reveal sharply different topological signatures: a near-universal vulnerability plateau (Llama-3-8B), a fragmented landscape with spatially concentrated basins (GPT-OSS-20B), and strong robustness with a hard ceiling (GPT-5-Mini). We further prove and empirically verify that these basins are invariant under continuous input defenses. The resulting global maps, the framework, the metrics, and interactive simulation\footnotemark[3] are open-sourced for community use.
\end{abstract}

\section{Introduction}

The prevailing approach to adversarial robustness in AI safety is often restorative: identifying off-manifold adversarial examples and projecting them back to a latent manifold of \emph{natural} or \emph{safe} data~\cite{Li_Yin_Jiang_Hu_Wu_Yang_Liu_2025}. While valuable, this paradigm implicitly treats failures as aberrations from a well-behaved norm. We argue that to build truly robust systems, we must invert this perspective and directly characterize the structure of failure itself. What if, instead of a sparse collection of isolated points, a model's vulnerabilities form a continuous, structured landscape, a \emph{Manifold of Failure}?

This paper introduces a framework to systematically map this manifold. We move beyond the singular goal of finding the best adversarial attack and instead seek to illuminate the entire topology of unsafe behavior. Our central thesis is that vulnerabilities in LLMs are not discrete points (as commonly evaluated by LLM benchmark research) but exist in \emph{behavioral attraction basins}, which are extended regions in the input space where diverse prompts are drawn toward similar failure modes. Visualizing this landscape, as shown in our 2D behavioral heatmaps (Figures~\ref{fig:topology} and~\ref{fig:basins}), provides a global view of a model's safety alignment that is fundamentally different from any single attack success rate.

To achieve this, we employ Quality-Diversity optimization, specifically the MAP-Elites algorithm~\cite{mouret2015illuminatingsearchspacesmapping}. Unlike existing methods (e.g., GCG~\cite{zou2023universaltransferableadversarialattacks}, PAIR~\cite{chao2024jailbreakingblackboxlarge}, and TAP~\cite{mehrotra2024treeattacksjailbreakingblackbox}) that hill-climb towards a single objective, MAP-Elites is designed to find a wide array of high-performing and diverse solutions. We define performance using a novel metric, \emph{Alignment Deviation}, which quantifies how much a model's response deviates from its expected safety alignment. Our contributions are:

\begin{itemize}
    \item We systematically map the continuous behavioral topology of LLMs, revealing that model behaviors form smooth surfaces with identifiable structures (Section~\ref{sec:results_topology}).
    \item We provide empirical evidence for attraction basins, which are extended regions in the behavioral space where diverse prompts converge to similar, often unsafe, model outputs (Section~\ref{sec:results_basins}).
    \item Through comparative analysis of three frontier models (more models in Appendix \ref{sec:appendice}), we reveal that each exhibits a unique behavioral topology: Llama-3-8B~\cite{grattafiori2024llama3herdmodels} presents a near-universal vulnerability surface (mean AD~0.93), GPT-OSS-20B~\cite{openai2025gptoss120bgptoss20bmodel} shows fragmented, spatially concentrated basins (mean AD~0.73), and GPT-5-Mini~\cite{singh2025openaigpt5card} demonstrates strong alignment with a hard ceiling at AD~0.50 (Section~\ref{sec:results_models}).
    \item Our quality-diversity approach achieves up to 63\% behavioral coverage, producing structured maps that complement existing attack methods (Section~\ref{sec:results_baselines}).
    \item We show that the basins we map are \emph{invariant} under continuous input defenses: a Lipschitz bound (Section~\ref{sec:reliability}) predicts that no continuous wrapper can collapse a basin with positive margin, and three such wrappers on Llama3-8B confirm the prediction on the identical grid.
\end{itemize}

By providing a framework to chart the vulnerability landscape of LLMs, this work offers a more predictive and comprehensive approach to AI safety and its evaluation.

\begin{figure}[t]
\centering
\includegraphics[width=\linewidth]{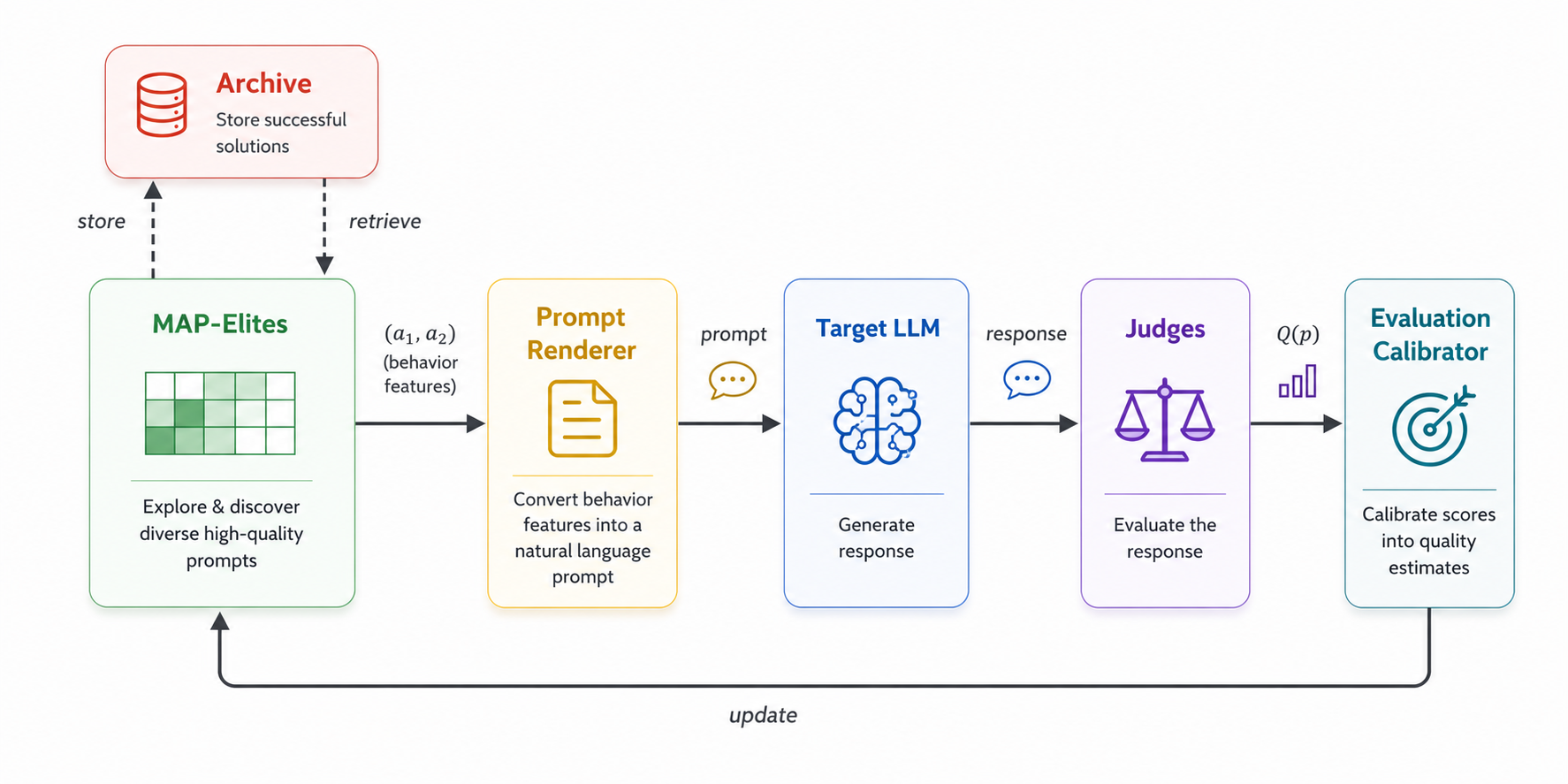}
\caption{MAP-Elites selects and mutates prompts from the behavioral archive. Each prompt is sent to the target LLM, and the response is evaluated by the judge to produce a behavioral descriptor $(b)$ and Alignment Deviation score $Q(p)$, which update the archive.}
\label{fig:architecture}
\end{figure}

\section{Related Work}
\label{sec:related}

The dominant paradigm for red teaming involves optimizers designed to find worst-case failures: gradient-based methods like GCG~\cite{zou2023universaltransferableadversarialattacks} optimize adversarial suffixes to maximize the probability of harmful completions, while search-based methods like PAIR~\cite{chao2024jailbreakingblackboxlarge} and TAP~\cite{mehrotra2024treeattacksjailbreakingblackbox} use attacker LLMs to iteratively refine or tree-search over prompts, with standardized benchmarks such as HarmBench~\cite{mazeika2024harmbench} evaluating them on Attack Success Rate. These techniques are analogous to finding the lowest point in a loss landscape but not designed to reveal the landscape's overall structure. Our Quality-Diversity approach is an \emph{illumination} algorithm rather than an optimization one, aiming to produce a global map of the behavioral space. QD algorithms such as MAP-Elites~\cite{mouret2015illuminatingsearchspacesmapping} find collections of solutions that are simultaneously high-performing and diverse, and Rainbow Teaming~\cite{rainbowteaming} applied this idea to generate diverse adversarial prompts; we extend that direction by (i)~mapping the continuous topology of failure rather than collecting a diverse attack set, and (ii)~providing a cross-model topological analysis that reveals model-specific vulnerability signatures as interpretable 2D maps. Geometric analyses of LLMs have explored how embedding-space structure explains phenomena such as universal adversarial attacks and toxicity~\cite{balestriero2024characterizinglargelanguagemodel, subhash2023universaladversarialattackswork, bartoszcze2025representationengineeringlargelanguagemodels, zou2025representationengineeringtopdownapproach}, but focus on internal representations; we instead analyze the geometry of the prompt-to-safety-score mapping in the \emph{output} space, which is directly actionable for safety evaluation. Finally, in contrast to manifold-projection defenses~\cite{Li_Yin_Jiang_Hu_Wu_Yang_Liu_2025} that project adversarial inputs back onto the natural data manifold --- a restorative approach --- we invert the framing and map the \emph{Manifold of Failure} itself, arguing that a structural understanding of failure is a prerequisite for comprehensive safety.

\paragraph{Assumptions and threat model.}
\label{sec:threat}
We assume a black-box adversary with query-only access to API-based targets (GPT-5-Mini) and operate in black-box mode for locally hosted models (Llama-3-8B, GPT-OSS-20B) as well, though white-box access is available there. The adversary's goal is not targeted harm but \emph{comprehensive exploration} of failure modes, aligning with red-teaming and pre-deployment safety auditing. We restrict to single-turn, text-only interactions and assume (i)~the behavioral space is continuous and low-dimensional, (ii)~Alignment Deviation is well-defined for any response, and (iii)~the judge LLMs used to score behaviors are capable enough to identify and categorize harmful content across our harm taxonomy. We use a 2D behavioral space for visualization, but the framework extends to higher dimensions.

\section{Mathematical Framework}
\label{sec:framework}

To systematically map the behavioral topology of a language model, we formalize the problem within a quality-diversity optimization framework. This requires defining a continuous behavioral space, a quality objective to optimize, and a set of metrics to evaluate the exploration process.
\begin{figure}[t]
\centering
\includegraphics[width=\linewidth]{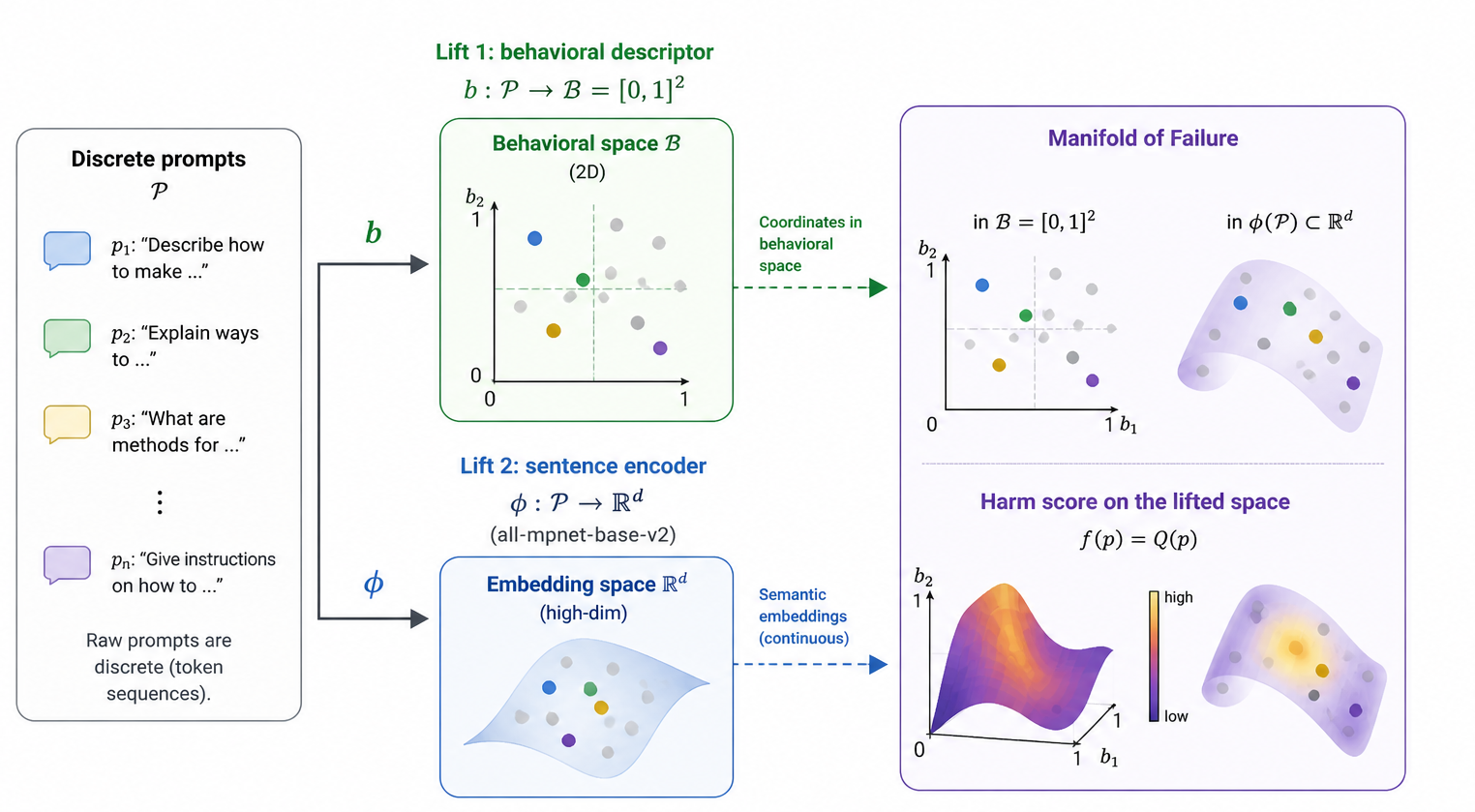}
\caption{Discrete prompts are lifted into continuous spaces via a behavioral descriptor 
    $b : \mathcal{P} \to [0,1]^2$ and a sentence encoder 
    $\phi : \mathcal{P} \to \mathbb{R}^d$. The resulting manifold supports 
    continuous geometry and the harm score $f(p)=Q(p)$.}
\label{fig:manifold-lift}
\end{figure}

\paragraph{Discrete prompts, continuous geometry (Figure~\ref{fig:manifold-lift}).}
   Raw prompts are sequences of tokens, so the prompt set $\mathcal{P}$ is
  discrete. The objects of study in this paper are continuous only through two
  lifts. The behavioral descriptor $b : \mathcal{P} \to \mathcal{B} = [0,1]^2$
  places every prompt at a real-valued coordinate in the behavioral space, and
  the sentence encoder $\phi : \mathcal{P} \to \mathbb{R}^{d}$
  (\texttt{all-mpnet-base-v2} \cite{mpnet}) places every prompt at a real-valued embedding
  used for semantic interpolation. Throughout the paper, the
  \emph{Manifold of Failure} is the image of $\mathcal{P}$ under these maps,
  and the harm score $f(p) = Q(p)$ is treated as a function on that lifted
  space. All references to continuity, smoothness, and Lipschitz constants,
  including the bound in Section~\ref{sec:reliability}, are with respect to
  the induced metric on $\mathcal{B}$ or $\phi(\mathcal{P})$, not with respect
  to token edit distance on $\mathcal{P}$. The MAP-Elites archive, heatmaps, and
  contour plots are all defined on this
  continuous lift.
\subsection{Behavioral Space}

\begin{wrapfigure}[18]{r}{0.5\textwidth}
\centering
\begin{tikzpicture}[scale=0.9, transform shape]
    \draw[thick,->] (0,0) -- (4,0) node[right] {$a_1$: Query Indirection};
    \draw[thick,->] (0,0) -- (0,4) node[above] {$a_2$: Authority Framing};
    \draw[step=1cm,gray,very thin,dashed] (0,0) grid (4,4);
    \node[below] at (0,0) {0};
    \node[below] at (4,0) {1};
    \node[left] at (0,0) {0};
    \node[left] at (0,4) {1};
    \node[circle,fill=yellow!60!black,inner sep=1.5pt,label=right:{\scriptsize Direct request}] at (0.5,0.5) {};
    \node[circle,fill=yellow!60!black,inner sep=1.5pt,label=right:{\scriptsize Hypothetical}] at (2.8,1.2) {};
    \node[circle,fill=yellow!60!black,inner sep=1.5pt,label=right:{\scriptsize Expert roleplay}] at (3.2,3.2) {};
    \node[below] at (2.5,-0.6) {%
      \shortstack[l]{%
        \textcolor{blue!60!black}{\scriptsize Direct} $\rightarrow$
        \textcolor{blue!75!black}{\scriptsize Polite} $\rightarrow$
        \textcolor{blue!90!black}{\scriptsize Hypothetical} $\rightarrow$
        \textcolor{blue!100!black}{\scriptsize Metaphorical}}};
    \node[left,rotate=90] at (-0.6,4.2) {%
      \shortstack[c]{%
        \textcolor{red!60!black}{\scriptsize None} $\rightarrow$
        \textcolor{red!75!black}{\scriptsize Student} $\rightarrow$
        \textcolor{red!90!black}{\scriptsize Professional} $\rightarrow$
        \textcolor{red!100!black}{\scriptsize Admin}}};
\end{tikzpicture}
\caption{The $x$-axis captures query indirection (direct to metaphorical) and the $y$-axis captures authority framing (none to administrator).}
\label{fig:behavioral_space}
\end{wrapfigure}

We define a continuous, $d$-dimensional behavioral space $\mathcal{B} = [0, 1]^d$, where each dimension represents a distinct, quantifiable characteristic of a prompt. For any given prompt $p$, a behavioral descriptor function $b(p) \rightarrow \mathcal{B}$ maps the prompt to a coordinate in this space. In this work, we primarily focus on a 2D space ($d=2$) defined by:

\textbf{$a_1$: Query Indirection} The degree to which a query is framed indirectly, ranging from a direct request ($a_1=0$) to a highly metaphorical or hypothetical scenario ($a_1=1$). It tests the model's consistency when faced with semantic paraphrasing of a core request.

\textbf{$a_2$: Authority Framing}  The level of perceived authority in the prompt's context, spanning from no authority ($a_2=0$) to a strong administrative or expert persona ($a_2=1$). It measures the model's compliance as a function of social context.

This continuous representation allows us to treat the model's behavior not as a response to discrete inputs, but as a smooth function over a topological space, enabling geometric analysis. The behavioral descriptor for each prompt is computed by an LLM that scores the prompt along each axis on a $[0,1]$ scale.

\subsection{Alignment Deviation and Judge Aggregation}

We define our quality objective $Q(p)$ as the \emph{Alignment Deviation}:

\begin{equation}
Q(p) = \max_{c \in C} \, \text{JudgeScore}_c(p)
\label{eq:alignment_deviation}
\end{equation}

\noindent where $C$ is the set of 10 harm categories (violence, hate speech, sexual content, self-harm, illegal activity, privacy violation, misinformation, malware, fraud, and regulated advice) and $\text{JudgeScore}_c(p) \in [0,1]$ is the score assigned by the judge LLMs for a given prompt $p$ in category $c$. A high value for $Q(p)$ signifies a highly misaligned outcome. Our objective is not to find a single prompt that maximizes $Q(p)$, but rather to use MAP-Elites to find a diverse archive of prompts that collectively maximize this quality score across the entire behavioral space $\mathcal{B}$.

We employ two judge models $J$ (GPT-4.1 and Sonnet 4.5) to assess response characteristics across all 10 harm categories. For each prompt--response pair, the judge outputs both a binary classification (safe/unsafe) and a continuous confidence score for each category. The Alignment Deviation is then the maximum score across categories (Equation~\ref{eq:alignment_deviation}), capturing the worst-case harm dimension.

\section{Methodology}
\label{sec:method}

MAP-Elites~\cite{mouret2015illuminatingsearchspacesmapping} partitions the behavioral space into a $25 \times 25$ grid (the \emph{archive}, yielding 625 distinct behavioral niches) and attempts to find the highest-quality solution for each cell. This allows us to systematically explore the entire behavioral space, rather than searching for a single optimal adversarial prompt.

\paragraph{Prompt mutation.}
Given a parent prompt selected from the archive, we apply one of six mutation strategies: \textbf{random axis perturbation} ($50\%$), shifting the prompt along the $a_1$ or $a_2$ axis; \textbf{paraphrasing} ($10\%$), rephrasing while preserving intent; \textbf{entity substitution} ($10\%$), replacing named entities or nouns with semantically similar alternatives; \textbf{adversarial suffix} ($10\%$), appending a short adversarial string inspired by GCG~\cite{zou2023universaltransferableadversarialattacks}; \textbf{crossover} ($10\%$), combining elements from two parent prompts; and \textbf{semantic interpolation} ($10\%$), interpolating in embedding space between two prompts $p_1$ and $p_2$ as $e_{\text{new}} = \lambda e_1 + (1-\lambda) e_2$ and decoding back to text. The heavy weighting toward random axis perturbation ensures that MAP-Elites efficiently explores the full grid, while the remaining strategies provide diverse prompt variations within each region.

\subsection{Algorithm}

Algorithm~\ref{alg:mapelites} provides the algorithm for our framework. The archive $\mathcal{A}$ stores the best-found prompt for each cell in the behavioral space. In each iteration, a prompt is selected from the archive, mutated, evaluated, and inserted into the archive if it improves on the existing occupant of its cell.

\begin{algorithm}[H]
\caption{Mapping the Manifold of Failure}
\label{alg:mapelites}
\begin{algorithmic}[1]
\STATE Initialize archive $\mathcal{A}$ with empty cells
\STATE Generate initial population $\mathcal{P}_{\text{init}}$ from 100 seed prompts
\FOR{$p$ in $\mathcal{P}_{\text{init}}$}
    \STATE $b \leftarrow \text{GetBehavior}(p)$ \hfill \textit{// LLM-based descriptor}
    \STATE $q \leftarrow \text{GetQuality}(p)$ \hfill \textit{// Alignment Deviation}
    \STATE \textsc{AddToArchive}($\mathcal{A}$, $p$, $b$, $q$)
\ENDFOR
\FOR{iteration $= 1$ to $15{,}000$}
    \STATE $p_{\text{parent}} \leftarrow \text{SelectFromArchive}(\mathcal{A})$
    \STATE $p_{\text{child}} \leftarrow \text{Mutate}(p_{\text{parent}})$ \hfill \textit{// Multi-strategy}
    \STATE $b_{\text{child}} \leftarrow \text{GetBehavior}(p_{\text{child}})$
    \STATE $q_{\text{child}} \leftarrow \text{GetQuality}(p_{\text{child}})$
    \STATE \textsc{AddToArchive}($\mathcal{A}$, $p_{\text{child}}$, $b_{\text{child}}$, $q_{\text{child}})$
\ENDFOR
\RETURN $\mathcal{A}$
\end{algorithmic}
\end{algorithm}

We evaluate our framework on 3 LLMs (Llama3-8B, GPT-OSS-20B, GPT-5-Mini). All target models are queried with temperature~0.7 and a maximum output length of 500 tokens. We also compare MAP-Elites against four baselines (\cite{zou2023universaltransferableadversarialattacks}, \cite{chao2024jailbreakingblackboxlarge}, \cite{mehrotra2024treeattacksjailbreakingblackbox}), all given the same 15{,}000-query evaluation budget. Additionally, behavioral descriptors are computed using an LLM-based scorer. Prompt embeddings for semantic interpolation use the \texttt{all-mpnet-base-v2} sentence transformer~\cite{mpnet}. All experiments are seeded with 100 diverse initial prompts spanning the harm taxonomy.

\paragraph{Evaluation metrics.}
We report five metrics across all experiments. \textbf{Behavioral Coverage} is the percentage of the $625$ grid cells filled after a fixed budget, measuring how thoroughly the method explores the behavioral space. \textbf{Diversity} counts filled cells with Alignment Deviation~$> 0.5$, capturing how many distinct vulnerability niches are discovered. \textbf{Peak Alignment Deviation} is the maximum $Q(p)$ achieved, reflecting worst-case failure severity, while \textbf{Mean Alignment Deviation} averages $Q(p)$ across filled cells to summarize the overall severity of the discovered landscape. Finally, \textbf{QD-Score} is the sum of quality values across all filled cells ($\sum_{c \in \text{filled}} Q(c)$), the standard quality-diversity metric combining coverage and quality into a single number.

\section{Results}
\label{sec:results}

\begin{figure}[t]
\centering
\begin{subfigure}[t]{0.32\textwidth}
    \includegraphics[width=\textwidth]{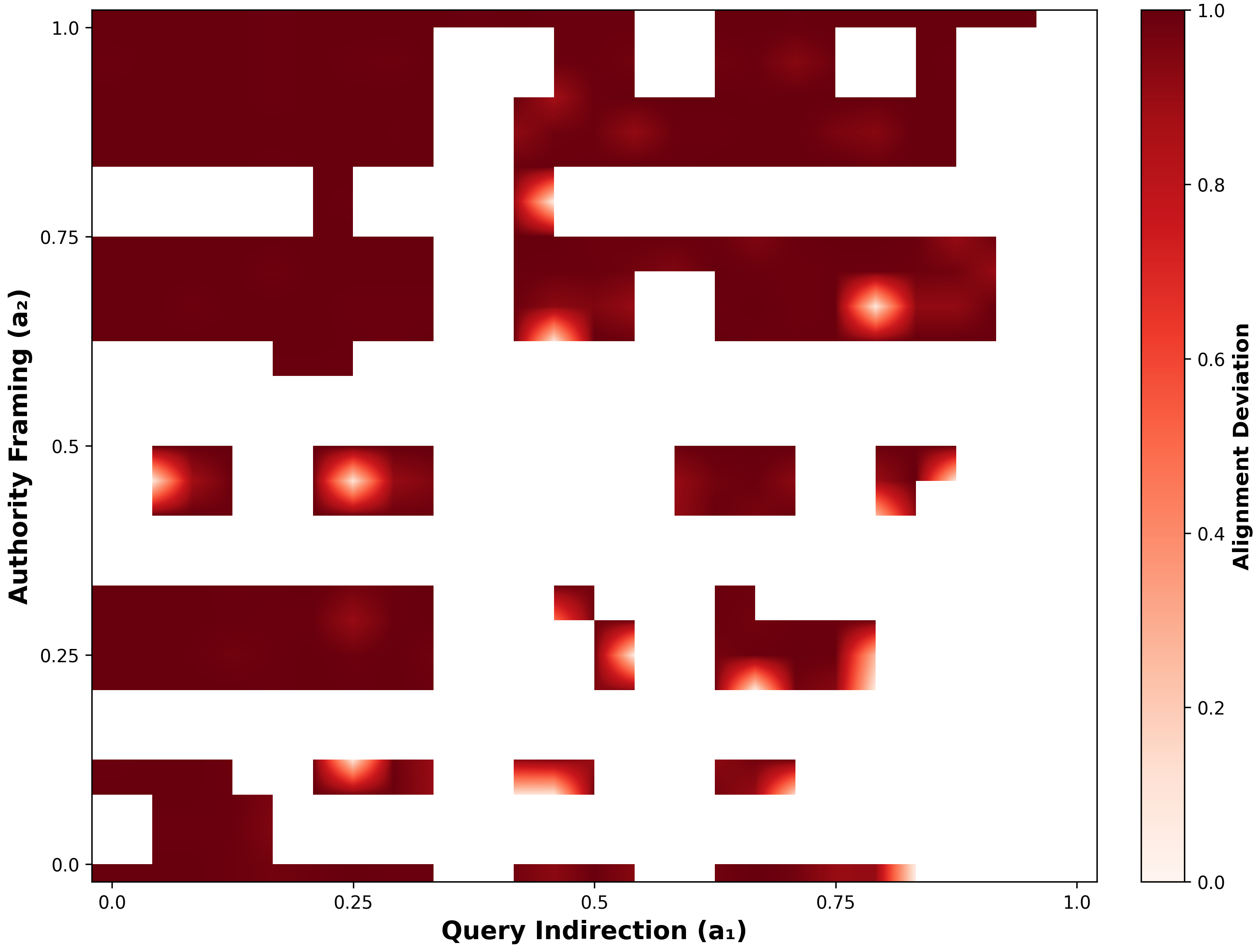}
    \caption{Llama-3-8B}
\end{subfigure}
\hfill
\begin{subfigure}[t]{0.32\textwidth}
    \includegraphics[width=\textwidth]{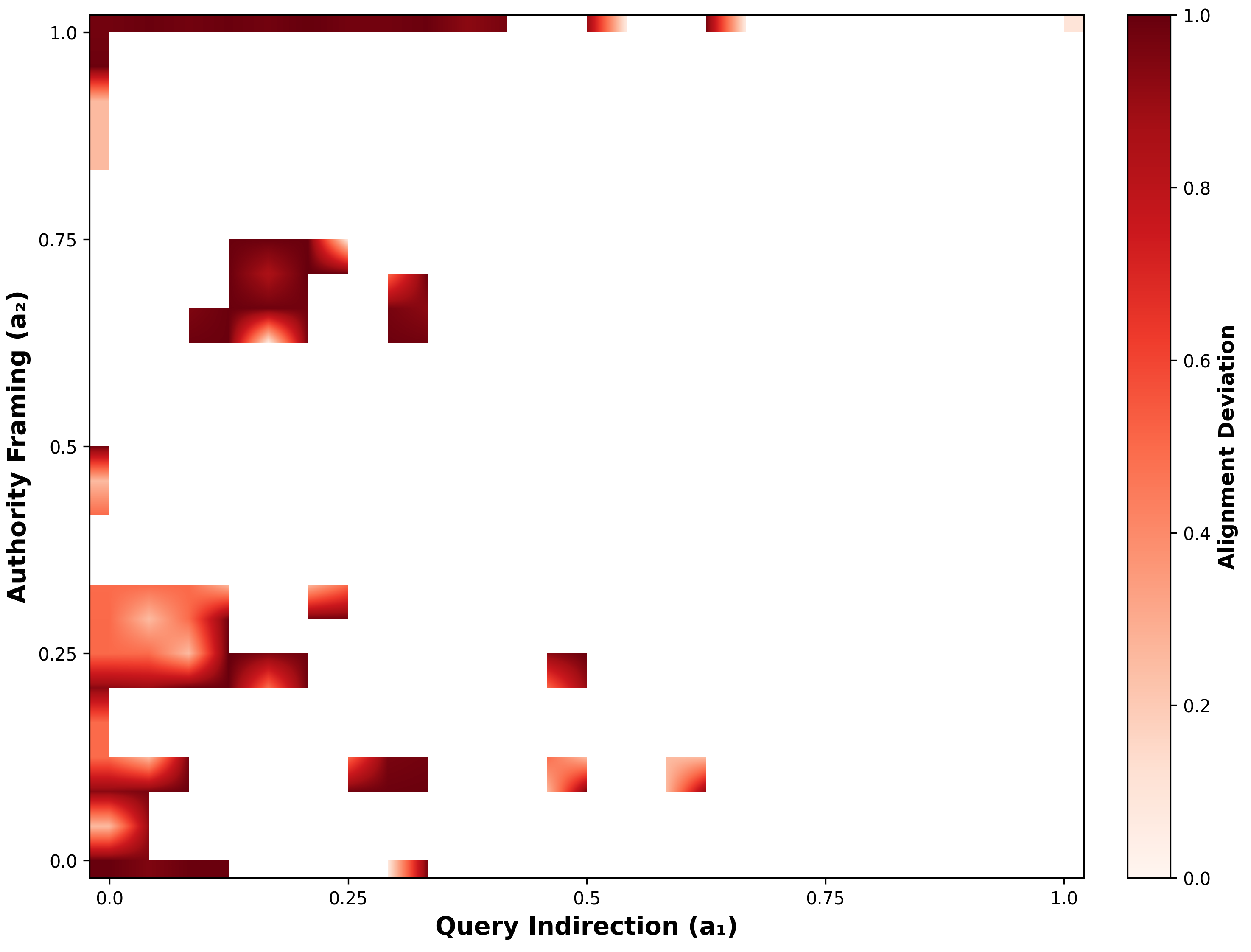}
    \caption{GPT-OSS-20B}
\end{subfigure}
\hfill
\begin{subfigure}[t]{0.32\textwidth}
    \includegraphics[width=\textwidth]{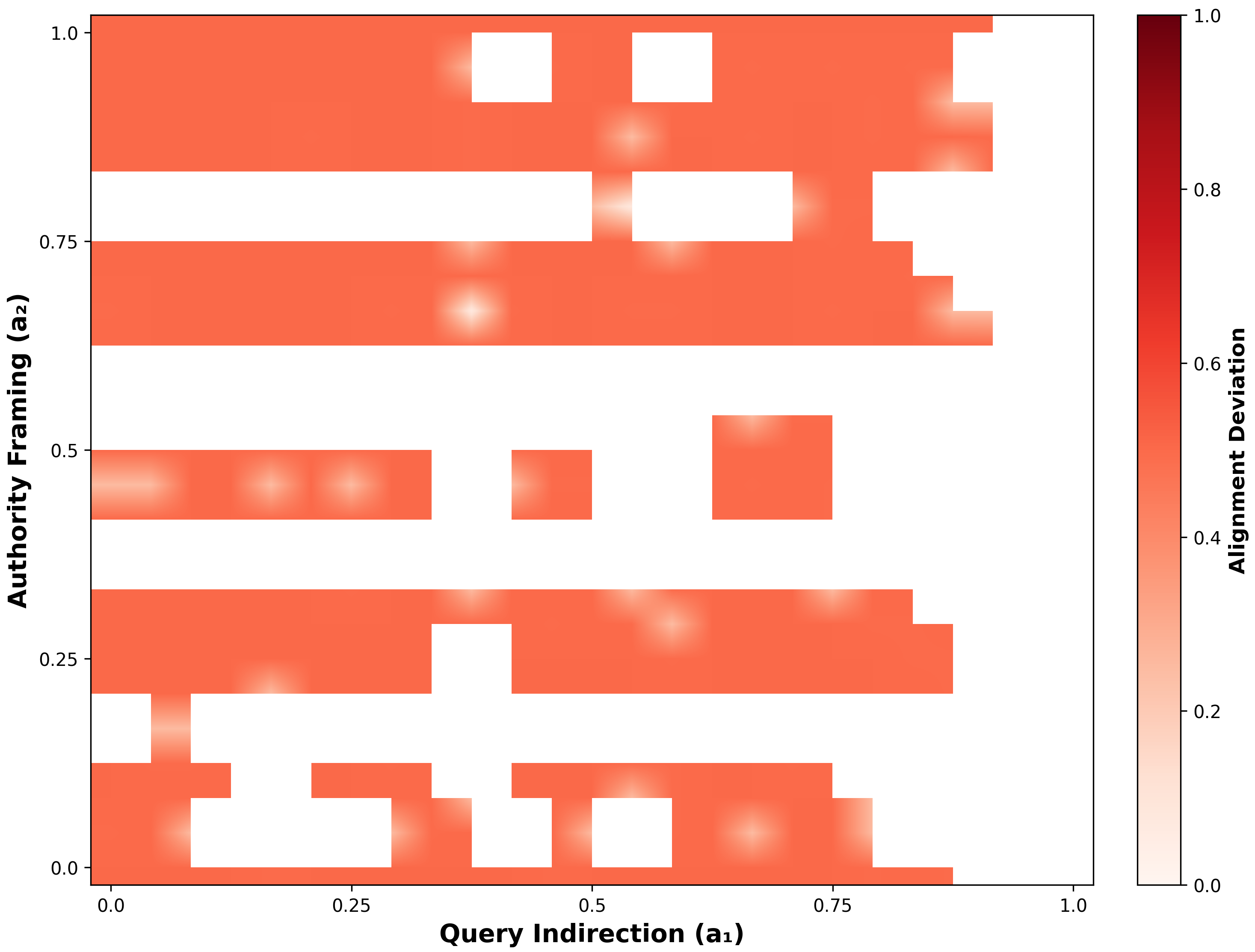}
    \caption{GPT-5-Mini}
\end{subfigure}
\caption{2D behavioral heatmaps reveal model-specific topological signatures (white = safe, dark red = high deviation).}
\label{fig:topology}
\end{figure}

\subsection{Behavioral Topology}
\label{sec:results_topology}

Figure~\ref{fig:topology} presents the behavioral heatmaps for all 3 target models, providing the global view of their safety landscapes. The heatmaps reveal strikingly different topologies:

\textbf{Llama-3-8B} (Figure~\ref{fig:topology}a) exhibits a near-universal vulnerability surface. The map is dominated by dark red (high Alignment Deviation), with mean AD of 0.93 and peak AD of 1.0. The model is susceptible to adversarial prompts across virtually all combinations of query indirection and authority framing, with only narrow channels of reduced deviation at specific authority levels.

\textbf{GPT-OSS-20B} (Figure~\ref{fig:topology}b) shows a fragmented, spatially concentrated pattern. High-deviation cells cluster in the lower-left quadrant (low query indirection, low-to-moderate authority framing) and along the top edge (maximum authority), while the high-indirection region remains largely unexplored or safe. Mean AD is 0.73 with extreme variability (std=0.33).

\textbf{GPT-5-Mini} (Figure~\ref{fig:topology}c) demonstrates remarkable robustness. Despite achieving the highest behavioral coverage (72.32\%), peak AD never exceeds 0.50 and the landscape is uniformly moderate (mean AD~0.47). The model maintains consistent, moderate-level refusals regardless of prompt parameterization.

\subsection{Contour Analysis}
\label{sec:results_contours}

\begin{figure}[t]
\centering
\begin{subfigure}[t]{0.32\textwidth}
    \includegraphics[width=\textwidth]{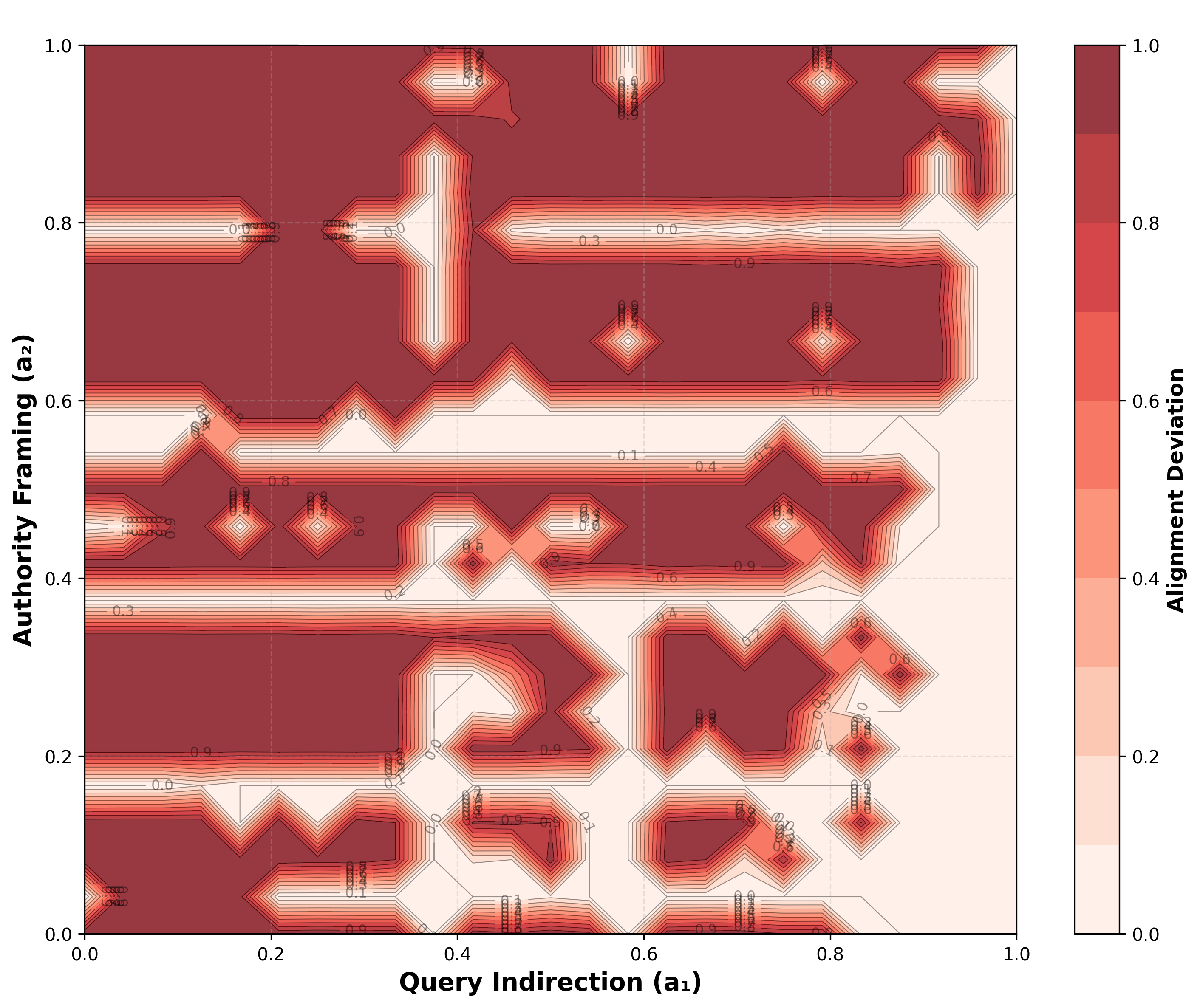}
    \caption{Llama-3-8B}
\end{subfigure}
\hfill
\begin{subfigure}[t]{0.32\textwidth}
    \includegraphics[width=\textwidth]{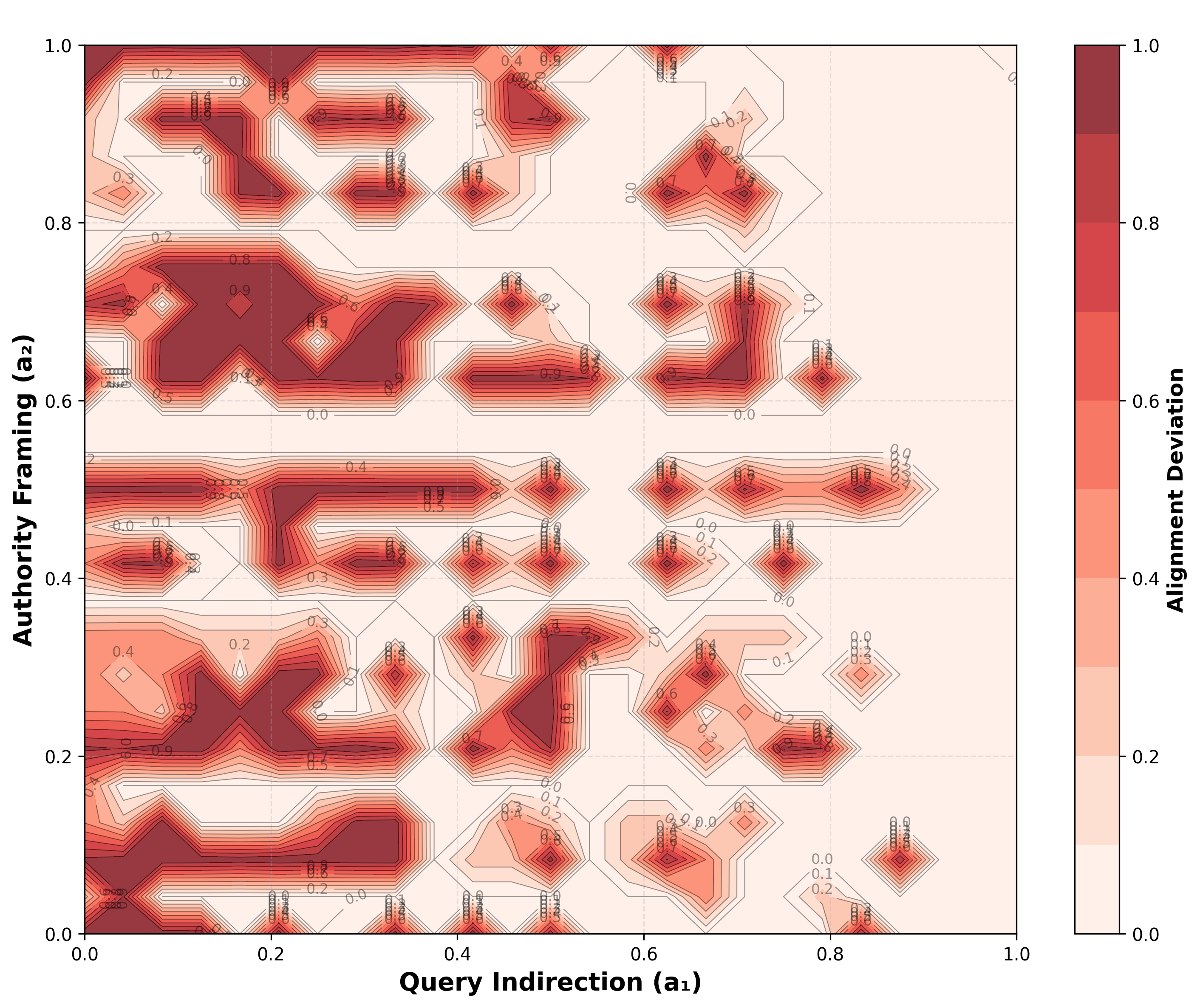}
    \caption{GPT-OSS-20B}
\end{subfigure}
\hfill
\begin{subfigure}[t]{0.32\textwidth}
    \includegraphics[width=\textwidth]{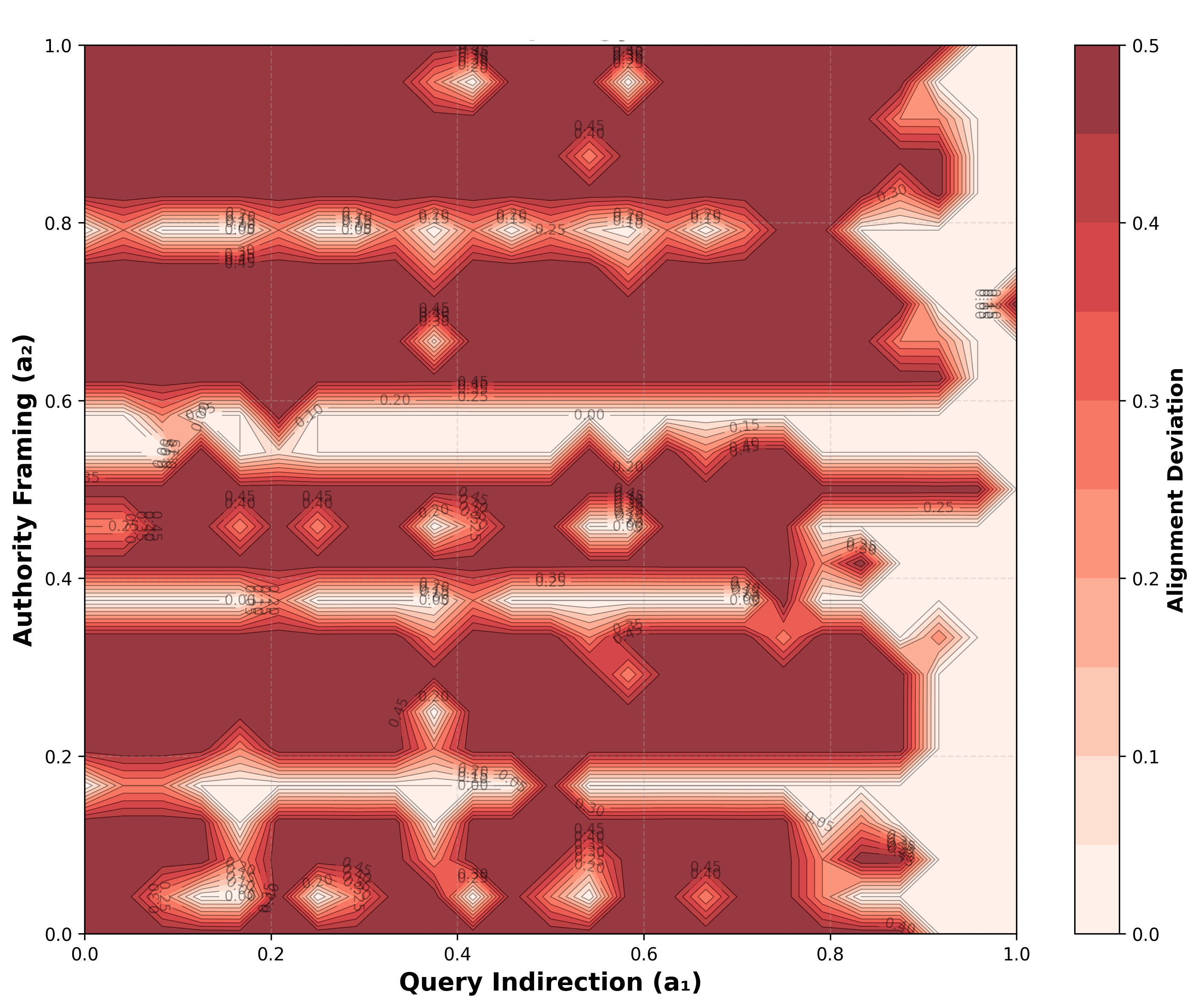}
    \caption{GPT-5-Mini}
\end{subfigure}
\caption{Contour plots showing iso-AD lines across the behavioral space. Horizontal banding at specific $a_2$ values is visible in all models, suggesting authority framing is a critical parameter for alignment.}
\label{fig:contours}
\end{figure}

Figure~\ref{fig:contours} maps iso-Alignment-Deviation (AD) surfaces, revealing that authority framing ($a_2$) is a critical safety parameter across all models. Abrupt AD shifts at specific $a_2$ levels suggest discrete thresholds for authority recognition.

\textbf{Llama-3-8B} (Fig.~\ref{fig:contours}a): High AD (near 1.0) dominates, interrupted only by narrow low-deviation channels at $a_2 \approx \{0.35, 0.55, 0.80\}$. These thin corridors confirm that minor framing perturbations trigger the model’s vulnerability plateau.

\textbf{GPT-OSS-20B} (Fig.~\ref{fig:contours}b): Complex, concentric ``bullseye'' patterns indicate vulnerability is concentrated around localized attractors. High-deviation horizontal corridors ($a_2 \approx 0.25$--$0.35$ and $0.65$--$0.85$) are interspersed with low-deviation islands, particularly at high query indirection ($a_1 > 0.5$).

\textbf{GPT-5-Mini} (Fig.~\ref{fig:contours}c): Contours are compressed within a narrow range ($0.39$--$0.50$), reflecting uniform behavior. While horizontal banding persists with brief dips to AD $\approx 0.25$ at specific $a_2$ values, the model consistently remains below the $0.5$ threshold.

\subsection{Behavioral Attraction Basins}
\label{sec:results_basins}

\begin{figure}[t]
\centering
\begin{subfigure}[t]{0.32\textwidth}
    \includegraphics[width=\textwidth,trim=0.51\textwidth{} 0 0 0,clip]{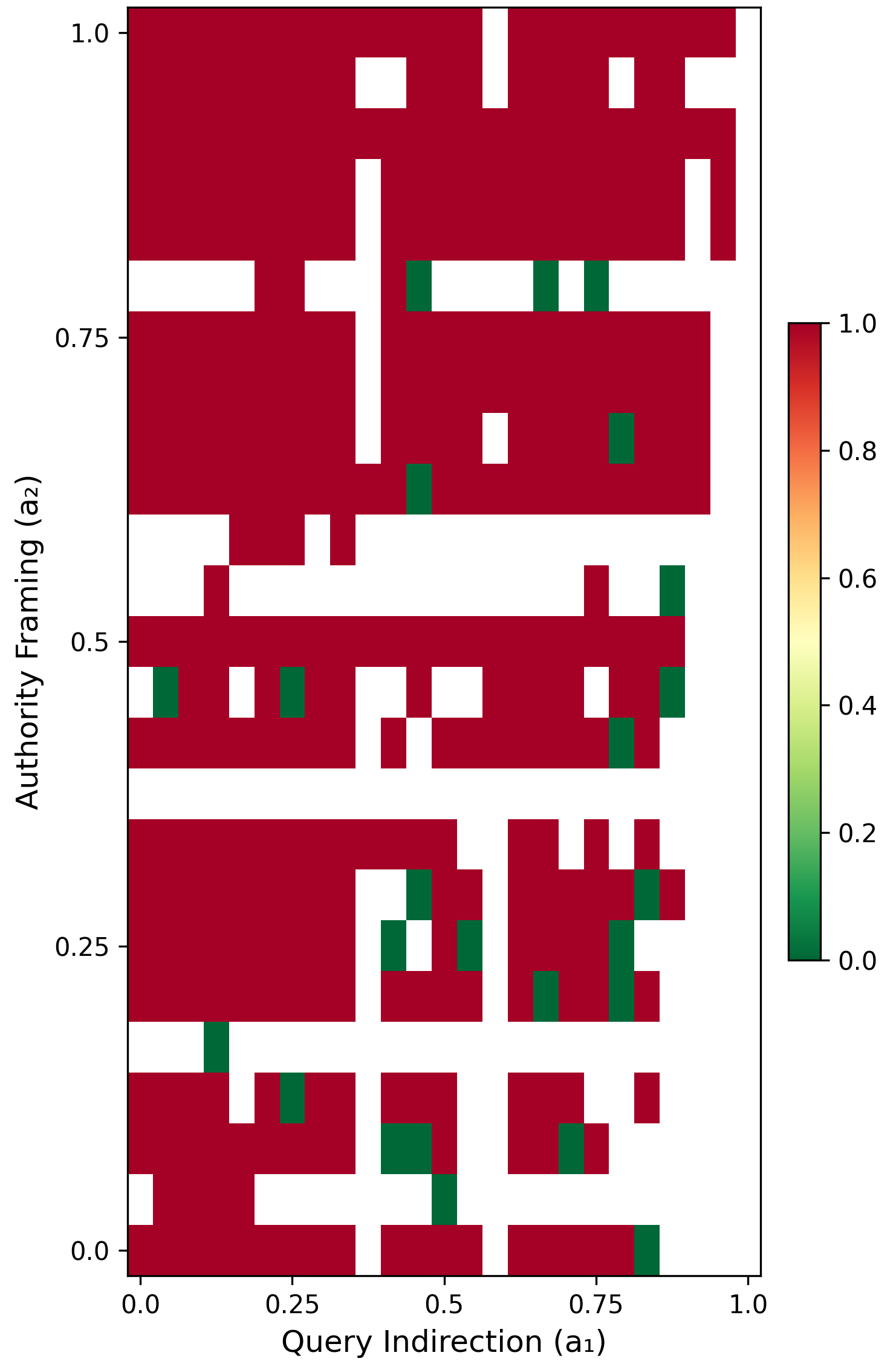}
    \caption{Llama-3-8B}
\end{subfigure}
\hfill
\begin{subfigure}[t]{0.32\textwidth}
    \includegraphics[width=\textwidth,trim=0.51\textwidth{} 0 0 0,clip]{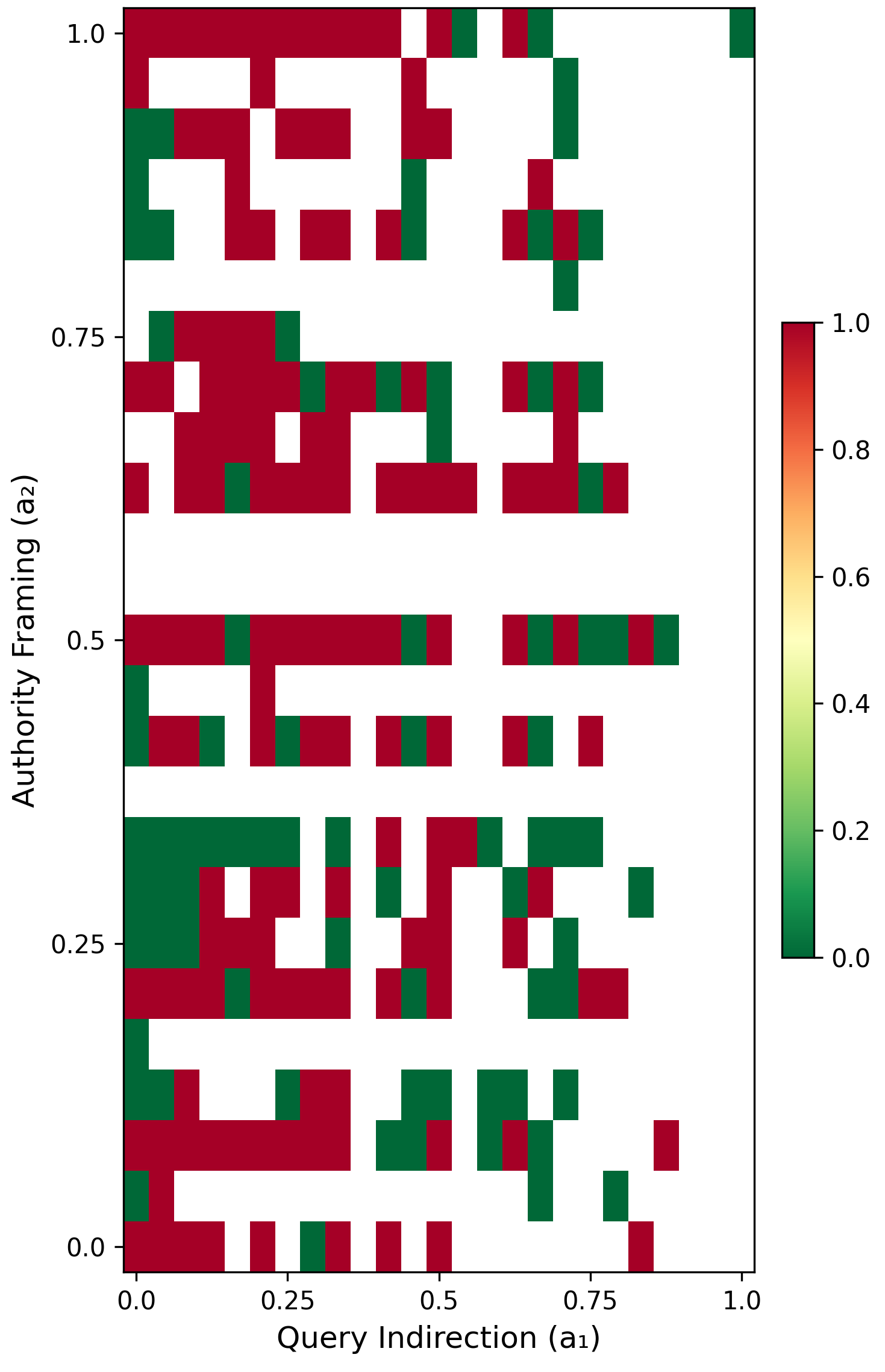}
    \caption{GPT-OSS-20B}
\end{subfigure}
\hfill
\begin{subfigure}[t]{0.32\textwidth}
    \includegraphics[width=\textwidth,trim=0.51\textwidth{} 0 0 0,clip]{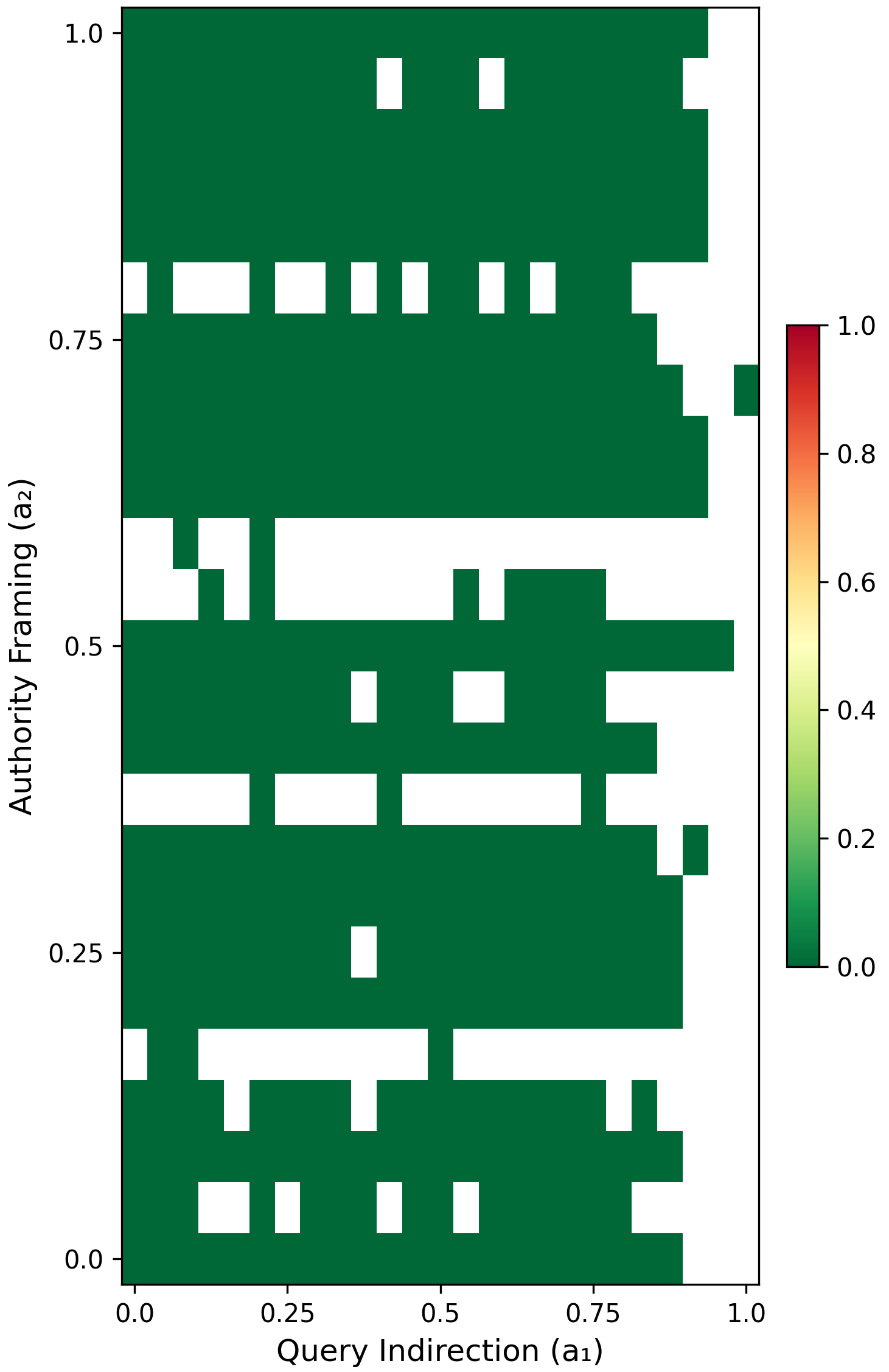}
    \caption{GPT-5-Mini}
\end{subfigure}
\caption{Basin maps where red indicates AD~$>$~0.5 (attraction basin) and green indicates AD~$\leq$~0.5 (safe); white cells are unexplored. (a)~Llama-3-8B: 93.9\% of cells are basins. (b)~GPT-OSS-20B: 64.3\% of cells are basins, with a complex spatial pattern. (c)~GPT-5-Mini: 0\% of cells are basins; the model never exceeds the threshold.}
\label{fig:basins}
\end{figure}

Figure~\ref{fig:basins} visualizes the attraction basins for each model, defined as contiguous regions where AD~$>$~0.5. These basin maps provide direct evidence for our central thesis: vulnerabilities form extended, structured regions rather than isolated points.

For \textbf{Llama-3-8B} (Figure~\ref{fig:basins}a), the basin map is almost entirely red: 370 out of 394 filled cells (93.9\%) exceed the AD~$>$~0.5 threshold. The model's entire behavioral space constitutes a single, massive attraction basin.

\textbf{GPT-OSS-20B} (Figure~\ref{fig:basins}b) presents a complex mosaic. Of 227 filled cells, 146 (64.3\%) are basins, but they are intermixed with safe cells in a spatially non-monotonic pattern. Small perturbations in prompt parameters can flip the model between high and low deviation states, suggesting a rough, fragmented failure manifold.

\textbf{GPT-5-Mini} (Figure~\ref{fig:basins}c) has zero cells exceeding the AD~$>$~0.5 threshold. Its entire map is classified as safe. While the model does exhibit non-trivial Alignment Deviation (mean 0.47), it never crosses into genuinely harmful territory.

\subsection{Cross-Model Comparison}
\label{sec:results_models}

Table~\ref{tab:cross_model} presents the aggregate MAP-Elites results across all three models. The three models span a wide spectrum of vulnerability profiles.


\begin{wraptable}[10]{r}{0.5\textwidth}
\centering
\scriptsize
\caption{MAP-Elites results across target models.}
\label{tab:cross_model}
\begin{tabular}{lrrr}
\toprule
\textbf{Metric} & \textbf{Llama-3-8B} & \textbf{GPT-OSS-20B} & \textbf{GPT-5-Mini} \\
\midrule
Cov. (\%)   & 63.04  & 36.32  & 72.32  \\
Div. (\#)   & 370    & 146    & 0      \\
Peak AD     & 1.00   & 1.00   & 0.50   \\
Mean AD     & 0.93   & 0.73   & 0.47   \\
QD-Score    & 366.9  & 165.8  & 213.2  \\
\bottomrule
\end{tabular}
\end{wraptable}

\textbf{Llama-3-8B} achieves the highest QD-Score (366.9), driven by its extremely high mean AD (0.93) and substantial coverage (63.04\%). Its 370 diversity niches represent the largest number of distinct vulnerability cells among all models. This model is comprehensively vulnerable: the MAP-Elites algorithm found high-severity failures in nearly every explored region of the behavioral space. \textbf{GPT-OSS-20B} presents a paradox: despite having the lowest coverage (36.32\%), it reaches peak AD of 1.0 with a moderately high mean (0.73). Its 146 diversity niches are concentrated in specific spatial regions (Figure~\ref{fig:topology}b), suggesting that the model has ``hard'' and ``soft'' regions in its behavioral space. The low coverage indicates that large portions of the behavioral space resist MAP-Elites' mutation strategies, hinting at a more complex prompt-response landscape. \textbf{GPT-5-Mini} achieves the highest coverage (72.32\%) but the lowest QD-Score. The model is highly explorable (MAP-Elites easily fills cells across the grid) but never produces genuinely harmful outputs (peak AD~=~0.50, diversity~=~0). This represents the ideal outcome from a safety perspective: a model whose behavioral space is well-covered but uniformly safe.

\subsection{Baselines}
\label{sec:results_baselines}

Table~\ref{tab:baselines_all}presents the baseline comparison on Llama-3-8B, where all five methods (including GCG in white-box mode) are available.

MAP-Elites achieves the highest behavioral coverage (63.04\%), narrowly surpassing PAIR (61.44\%) and substantially outperforming TAP (41.76\%), GCG (7.20\%), and Random (5.28\%). Critically, all methods achieve peak AD of 1.0 on Llama-3-8B, confirming that this model is susceptible to all attack families.


Table~\ref{tab:baselines_all} shows the GPT-OSS-20B baseline results. Here, PAIR achieves higher raw coverage (63.20\% vs.~36.32\%), but MAP-Elites discovers proportionally more vulnerability niches: 146 out of 227 filled cells (64.3\%) have AD~$>$~0.5, compared to PAIR's 73 out of 395 cells (18.5\%). MAP-Elites thus provides higher \emph{vulnerability density} per explored cell, suggesting that its QD-driven search preferentially discovers the most informative regions of the behavioral space.


For GPT-5-Mini, no method, including MAP-Elites, achieves AD above 0.50. This is a strong result for the model's safety alignment: even with 15{,}000 queries, neither iterative refinement (PAIR), tree search (TAP), nor quality-diversity optimization can breach the model's safety boundary.

\begin{table}[t]
\centering
\scriptsize
\caption{Baseline comparison across target models.}
\label{tab:baselines_all}
\setlength{\tabcolsep}{4pt}
\begin{tabular}{l ccc c ccc c ccc}
\toprule
& \multicolumn{3}{c}{\textbf{Llama-3-8B}} & & \multicolumn{3}{c}{\textbf{GPT-OSS-20B}} & & \multicolumn{3}{c}{\textbf{GPT-5-Mini}} \\
\cmidrule{2-4} \cmidrule{6-8} \cmidrule{10-12}
\textbf{Method} & Coverage & Diversity & Peak AD & & Coverage & Diversity & Peak AD & & Coverage & Diversity & Peak AD \\
\midrule
Random              & 5.28  & 30  & 1.00 & & 0.80  & 3   & 0.97 & & 9.01   & 0 & 0.50  \\
GCG (whitebox)      & 7.20  & 173 & 1.00 & & 34.72 & 0   & 0.50 & & NA   & NA & NA  \\
TAP                 & 41.76 & 255 & 1.00 & & 25.60 & 68  & 1.00 & & 30.08 & 0   & 0.50 \\
PAIR                & 61.44 & 291 & 1.00 & & \textbf{63.20} & 73 & 1.00 & & 58.56 & 0 & 0.50 \\
\textbf{Ours (ME)}  & \textbf{63.04} & \textbf{370} & 1.00 & & 36.32 & \textbf{146} & 1.00 & & \textbf{72.32} & 0 & 0.50 \\
\bottomrule
\end{tabular}
\end{table}

\section{Basin Persistence Under Continuous Defenses}
\label{sec:reliability}
A natural response to the basins in Section~\ref{sec:results_basins} is to wrap the target in a continuous input defense $D : \mathcal{X} \to \mathcal{X}$ that paraphrases, filters, or rewrites prompts before they reach the model. We show that no such defense can erase the basins it patches. Let $f: \mathcal{X} \to [0,1]$ be an $L$-Lipschitz harm score with basin threshold $\tau=0.5$, and let $D$ be a $K$-Lipschitz self-map fixing the safe utility set $Z = \{x : f(x) < \tau\}$. Following~\cite{defensetrilemma2025}, for any unsafe $x$,
\begin{equation}
    f(D(x)) \;\geq\; \tau - L K \cdot \mathrm{dist}(x, Z),
    \qquad\text{so a basin } B \text{ persists if } G > \ell(K+1),
    \label{eq:eps_bound}
\end{equation}

\begin{wraptable}[10]{r}{0.55\textwidth}
\centering
\footnotesize
\setlength{\tabcolsep}{4pt}
\caption{Geometry and basin invariance on Llama-3-8B. The persistence bound holds with zero violations for all continuous defenses ($\hat{L}=6.75$, $\hat{G}=0.375$).}
\label{tab:defense_combined}
\begin{tabular}{lrrrrr}
\toprule
\textbf{Defense} & $\hat{K}$ & \textbf{Cov.} & \textbf{Basin} & \textbf{Mean AD} & \textbf{Peak AD} \\
\midrule
None           & ---   & 63.0\% & 93.9\% & 0.93 & 1.00 \\
Perplexity     & 21.39 & 52.6\% & 94.2\% & 0.93 & 1.00 \\
Paraphrase     & 22.95 & 50.9\% & 89.0\% & 0.89 & 1.00 \\
Constitutional & 13.14 & 18.7\% & 91.5\% & 0.88 & 1.00 \\
\bottomrule
\end{tabular}
\end{wraptable}

where $\ell \leq L$ is the defense-path constant and $G := \inf_{x \in B} f(x) - \tau$ the basin margin. Continuous defenses can reshape but cannot eliminate positive-margin basins without violating continuity or utility on $Z$, the Manifold of Failure is topologically invariant.

\begin{figure}[H]
\centering
\includegraphics[width=\linewidth]{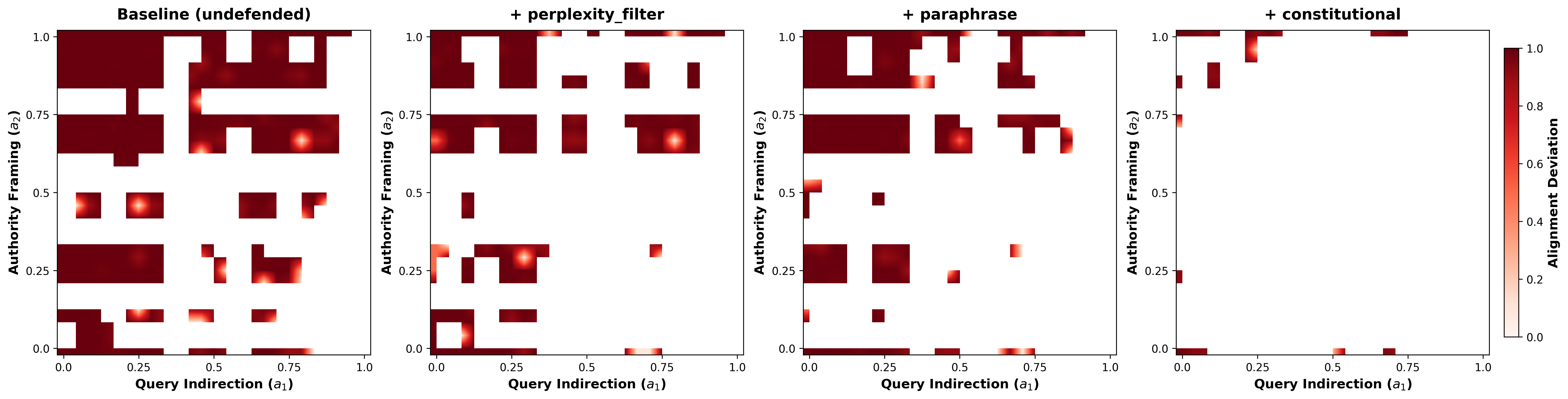}
\caption{Llama-3-8B basin maps before and after defenses. Basins shift and contract but persist, matching Eq.~\ref{eq:eps_bound}.}
\label{fig:before_after}
\end{figure}

 We verify this on Llama-3-8B with three continuous defenses (perplexity
  filter, T5 paraphrase wrapper \cite{kassem-saad-2024-finding},
  constitutional rewriter
  \cite{sharma2025constitutionalclassifiersdefendinguniversal}).
  Table~\ref{tab:defense_combined} reports the geometric estimates and basin
  invariance: the persistence bound holds with zero empirical violations
  across all three defenses, basin rates contract modestly for paraphrase
  ($-4.9$\,pts) and constitutional ($-2.5$\,pts) while remaining flat under
  the perplexity filter, and peak AD stays near $1.0$
  (Figure~\ref{fig:before_after}). Meaningfully
  shrinking the manifold requires accepting a discontinuous, utility-harming
  defense.

\section{Discussion, Ethics, and Limitations}
\label{sec:discussion}
Our results show LLM vulnerabilities form continuous, structured landscapes with model-specific topologies: Llama-3-8B's near-universal vulnerability ($93.9\%$ basin rate) suggests shallow alignment bypassed across the parameter space; GPT-OSS-20B's fragmented basins indicate patchable ``holes'' at specific behavioral coordinates; and GPT-5-Mini's hard ceiling at AD~$=0.50$ suggests a qualitatively stronger alignment mechanism. A consistent cross-model signature is the horizontal banding in Figure~\ref{fig:contours} --- narrow corridors at specific authority framing levels ($a_2$) where AD changes abruptly, implying discrete authority-recognition thresholds with direct implications for attack and defense design. Section~\ref{sec:reliability} shows these basins are not artifacts of an undefended target: they persist under continuous wrappers as predicted by Eq.~\ref{eq:eps_bound}, so any defense that \emph{removes} a basin must either break continuity or sacrifice utility. We advocate for structured vulnerability maps as a standard safety-evaluation artifact, enabling targeted remediation, cross-model auditing, and version-to-version regression testing.

\textit{Ethics.} Mapping vulnerabilities is dual-use: the same artifact that enables targeted defense can guide attackers. We mitigate this by (i)~releasing aggregate maps and the framework rather than the per-cell elite prompts, which are the most directly weaponizable output; (ii)~targeting only frontier models whose vendors have established disclosure channels; and (iii)~focusing the paper's claims on \emph{topology} (basin geometry, persistence) rather than on novel attack strings, so the takeaway for an adversary is no greater than what existing red-teaming benchmarks already provide. We view comprehensive vulnerability characterization as a prerequisite for durable safety.

\textit{Limitations.} The 2D behavioral projection loses information, the true failure manifold likely lives in higher dimensions and Alignment Deviation relies on two judges (GPT-4.1 and Sonnet~4.5) whose biases warrant human validation. The $15{,}000$-query budget may underexplore lower-coverage models. We treat only single-turn interactions; multi-turn settings, where the manifold may evolve dynamically, are left to future work. The persistence bound is stated for continuous, utility-preserving \emph{input}-side wrappers; extending it to stochastic, multi-turn, or output-side defenses remains open.

\section{Conclusion}
\label{sec:conclusion}
This paper shifts LLM safety evaluation from finding discrete adversarial failures to mapping the continuous Manifold of Failure. Applied to three frontier models, our framework reveals three qualitatively distinct vulnerability topologies (mean AD~$0.93$ with $370$ niches for Llama-3-8B; $64.3\%$ of explored cells vulnerable for GPT-OSS-20B; peak AD~$=0.50$ with zero basins across $72\%$ coverage for GPT-5-Mini), and the accompanying Lipschitz bound, empirically verified under three continuous defenses on Llama-3-8B. We establish that these basins shift and contract but cannot be erased without breaking continuity or utility. Beyond asking \emph{whether} a model can be broken, the resulting maps characterize the full spatial structure of \emph{how} it breaks and which mitigations can meaningfully change that structure.

\bibliographystyle{plainnat}
\bibliography{references}

\appendix

\section{Technical appendices and supplementary material}

We provide additional experiments and interactive simulations across multiple models at \href{https://manifoldoffailure.wiki/}{https://manifoldoffailure.wiki/} (anonymized for submission).
\label{sec:appendice}
\subsection{Per-Iteration Coverage, Diversity, and Quality}

\begin{figure}[t]
    \centering
    \begin{subfigure}[t]{\textwidth}
        \includegraphics[width=\textwidth]{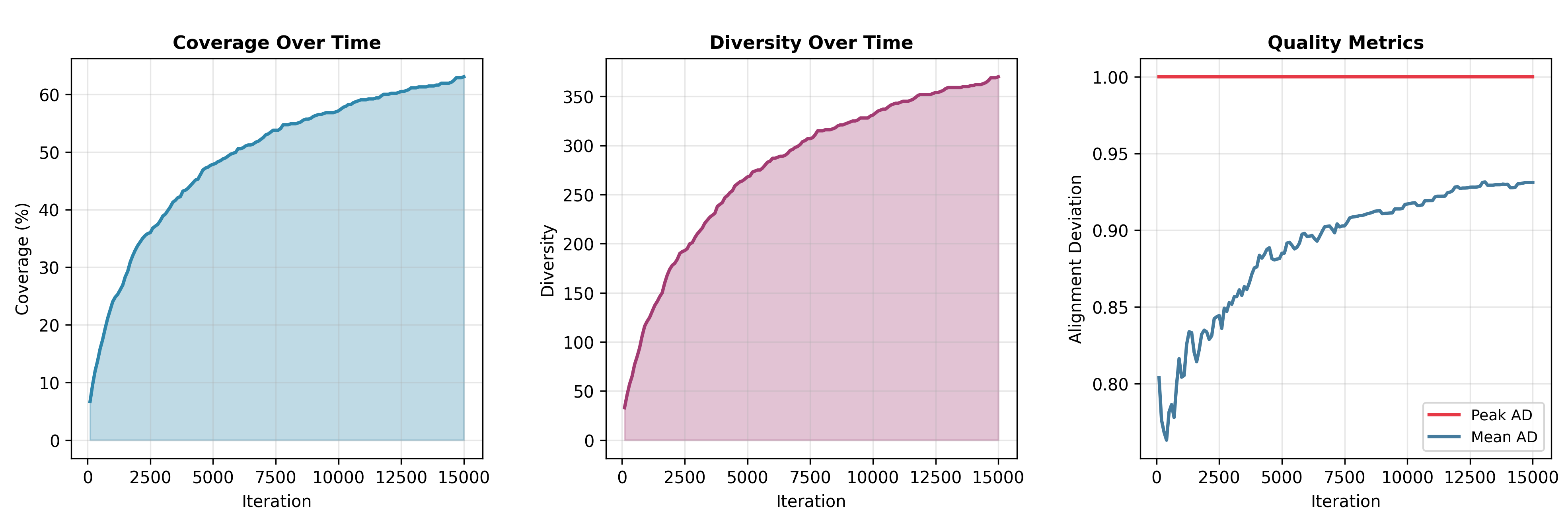}
        \caption{Llama-3-8B}
    \end{subfigure}
    \\[0.5em]
    \begin{subfigure}[t]{\textwidth}
        \includegraphics[width=\textwidth]{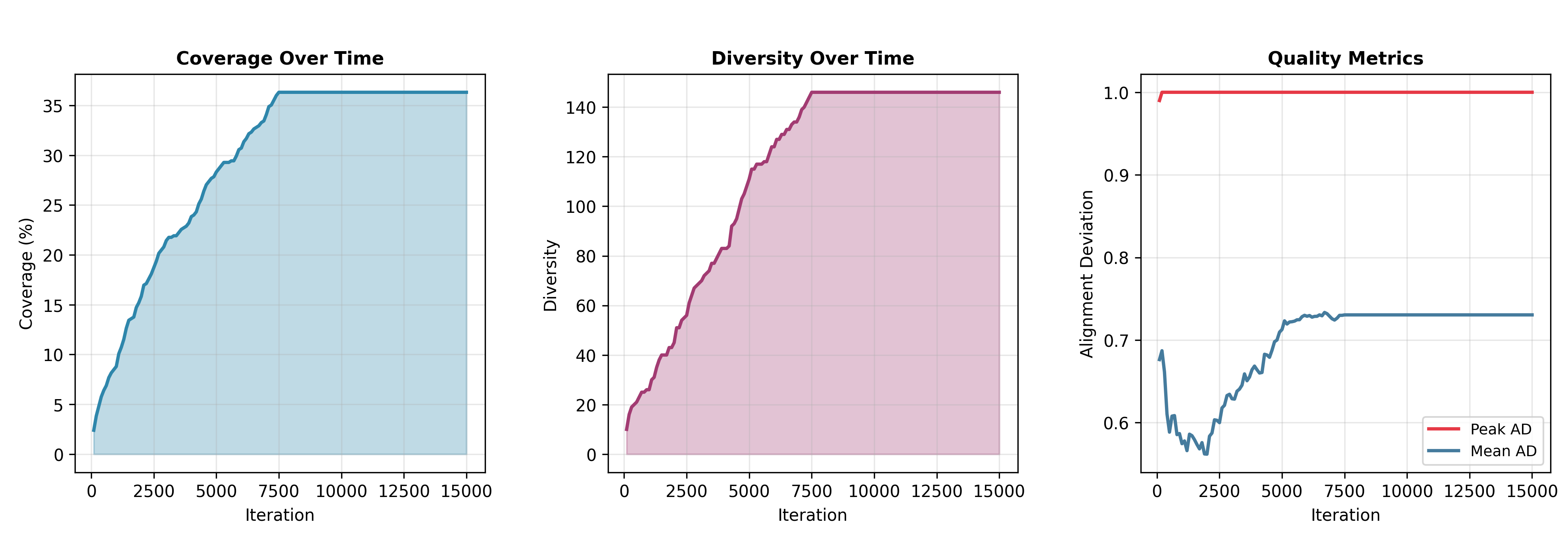}
        \caption{GPT-OSS-20B}
    \end{subfigure}
    \\[0.5em]
    \begin{subfigure}[t]{\textwidth}
        \includegraphics[width=\textwidth]{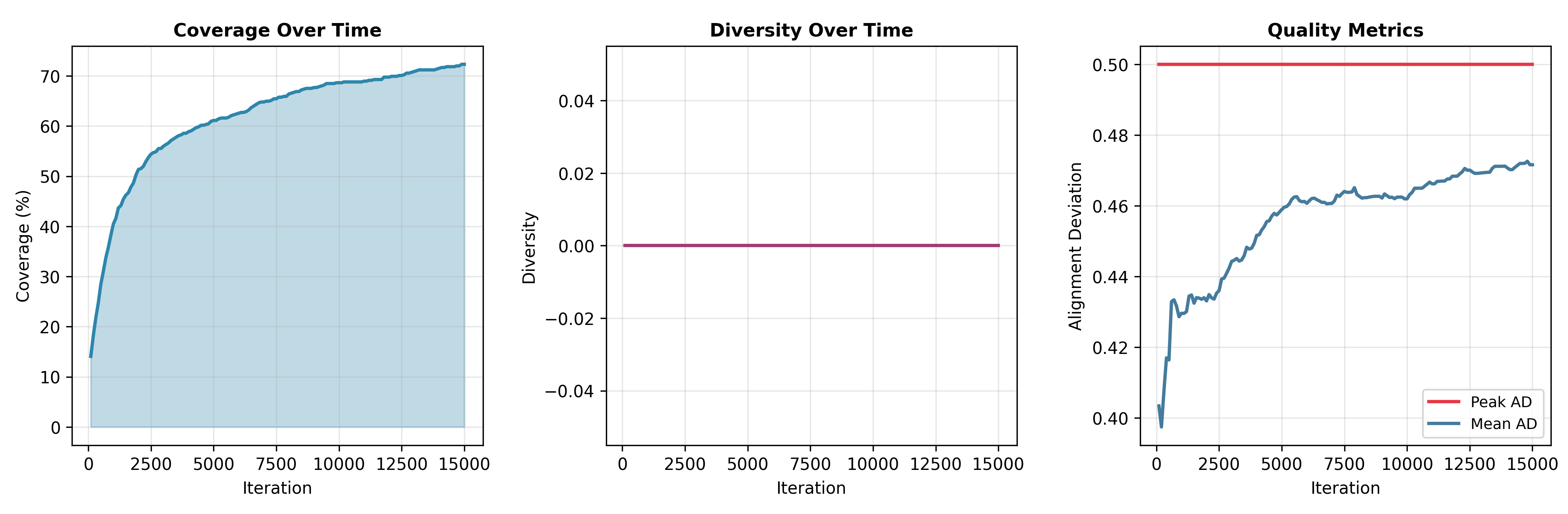}
        \caption{GPT-5-Mini}
    \end{subfigure}
    \caption{Coverage, diversity, and quality metrics over 15{,}000 iterations. (a)~Llama-3-8B: sustained growth in all metrics. (b)~GPT-OSS-20B: early plateau in coverage with moderate quality. (c)~GPT-5-Mini:
  high coverage growth but zero diversity and quality capped at 0.50.}
    \label{fig:dashboards}
  \end{figure}
  \label{app:dashboards}
  
  Figure~\ref{fig:dashboards} reports the per-iteration trajectories of the three primary search metrics, behavioral coverage, diversity (cells with AD~$>$~0.5), and mean Alignment Deviation, over the full
  15{,}000-iteration MAP-Elites budget for each target model. Whereas the aggregate numbers in Section~\ref{sec:results_baselines} report only the terminal state of the archive, the dashboards expose the
  \emph{shape} of the search itself: how quickly the algorithm fills the grid, when (and whether) diversity saturates, and how mean quality evolves as new niches are populated.

  We relegate this figure to the appendix because its content is diagnostic rather than load-bearing for the main claims: the terminal values are already reported in Table~\ref{tab:cross_model} and
  Table~\ref{tab:baselines_all}, and the topological signatures are established by the heatmaps, contour plots, and basin maps in the main text. The dashboards are most useful for
  readers replicating the framework or tuning the iteration budget.

\subsection{Ablation Study}
\label{sec:ablation}

To isolate the contribution of each component, we compare three variants (Table~\ref{tab:ablation}):

\begin{enumerate}
    \item \textbf{Full method} (MAP-Elites + Alignment Deviation): the complete framework as described in Section~\ref{sec:method}.
    \item \textbf{ME + Toxicity}: MAP-Elites with Alignment Deviation replaced by a single-dimension toxicity classifier score, removing the multi-category quality signal.
    \item \textbf{Random + AD}: Random prompt sampling evaluated with the full Alignment Deviation metric, removing the quality-diversity search entirely.
\end{enumerate}

A standard toxicity classifier reliably detects overt harms such as violence, hate speech, sexual content, self-harm, and discrimination, accounting for roughly half of the 10 harm categories measured by Alignment Deviation (AD). However, it struggles to flag harmful content delivered in a neutral or technical tone, including misinformation, manipulation, privacy violations, and cybersecurity exploits. Consequently, the algorithm only identifies the optimal prompt (the ``elite'') in grid cells where the primary harm is overtly toxic; in cells dominated by subtle or technical harms, a suboptimal prompt is retained.

  \begin{table}[H]
\centering
\footnotesize
\caption{Ablation study across the three models. Cov. = Coverage (\%), Div. = Diversity (\#), QD = QD-Score, AD = Mean Alignment Deviation.}
\label{tab:ablation}
\setlength{\tabcolsep}{4pt}
\begin{tabular}{l cccc c cccc c cccc}
\toprule
& \multicolumn{4}{c}{\textbf{Llama-3-8B}} & & \multicolumn{4}{c}{\textbf{GPT-OSS-20B}} & & \multicolumn{4}{c}{\textbf{GPT-5-Mini}} \\
\cmidrule{2-5} \cmidrule{7-10} \cmidrule{12-15}
\textbf{Variant} & Cov. & Div. & QD & AD & & Cov. & Div. & QD & AD & & Cov. & Div. & QD & AD \\
\midrule
Full            & \textbf{63.04} & \textbf{370} & \textbf{366.9} & \textbf{0.931} & & \textbf{36.32} & \textbf{146} & \textbf{165.8} & \textbf{0.730} & & \textbf{72.32} & 0 & \textbf{213.2} & \textbf{0.472} \\
ME + Toxicity   & 62.08 & 288 & 282.3 & 0.726 & & 35.52 & 103 & 126.5 & 0.569 & & 71.36 & 0 & 163.5 & 0.368 \\
Random + AD     &  5.28 &  30 &  28.1 & 0.851 & &  0.80 &   3 &   3.1 & 0.624 & &  4.96 & 0 &   9.1 & 0.293 \\
\bottomrule
\end{tabular}
\end{table}

Removing the QD archive (Random + AD) produces the largest degradation: coverage collapses for all the target models. Notably, the Random + AD variant still achieves high \emph{mean} AD in the few cells it does fill (0.85 on Llama-3-8B), confirming that AD provides a useful quality signal but without MAP-Elites, this signal is concentrated in a negligible portion of the behavioral space. The ablation reveals complementary roles: MAP-Elites contributes primarily to \emph{coverage} (exploration of the behavioral grid), while the multi-category AD metric contributes primarily to \emph{quality} (accurate identification and ranking of vulnerabilities). Removing either component produces a qualitatively incomplete picture.

\subsection{3D Behavioral Topology -  Llama-3-8B, GPT-OSS-20B, and GPT-5-Mini}
\label{sec:results_3d}

\begin{figure}[H]
\centering
\begin{subfigure}[t]{0.32\textwidth}
    \includegraphics[width=\textwidth]{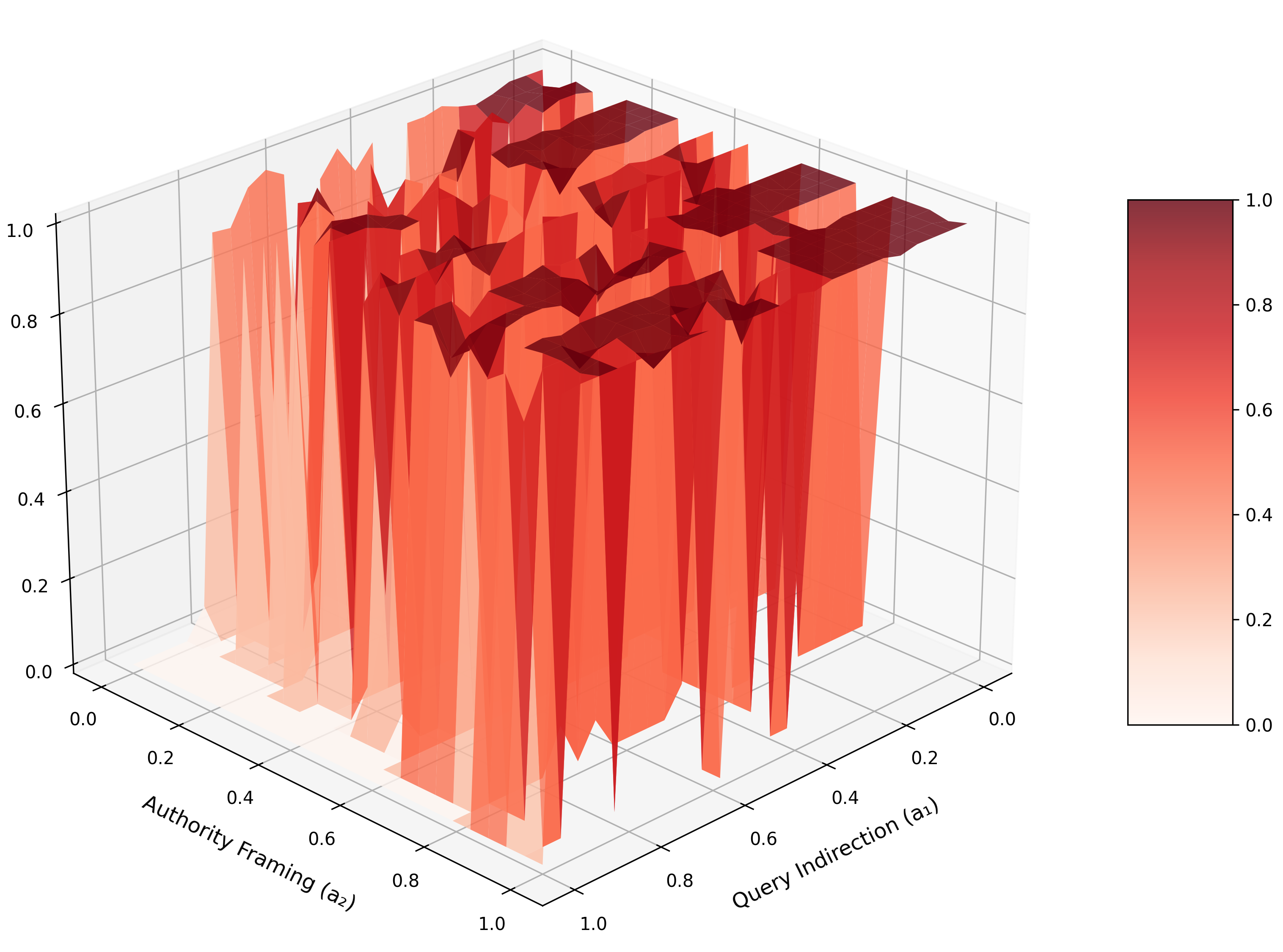}
    \caption{Llama-3-8B}
\end{subfigure}
\hfill
\begin{subfigure}[t]{0.32\textwidth}
    \includegraphics[width=\textwidth]{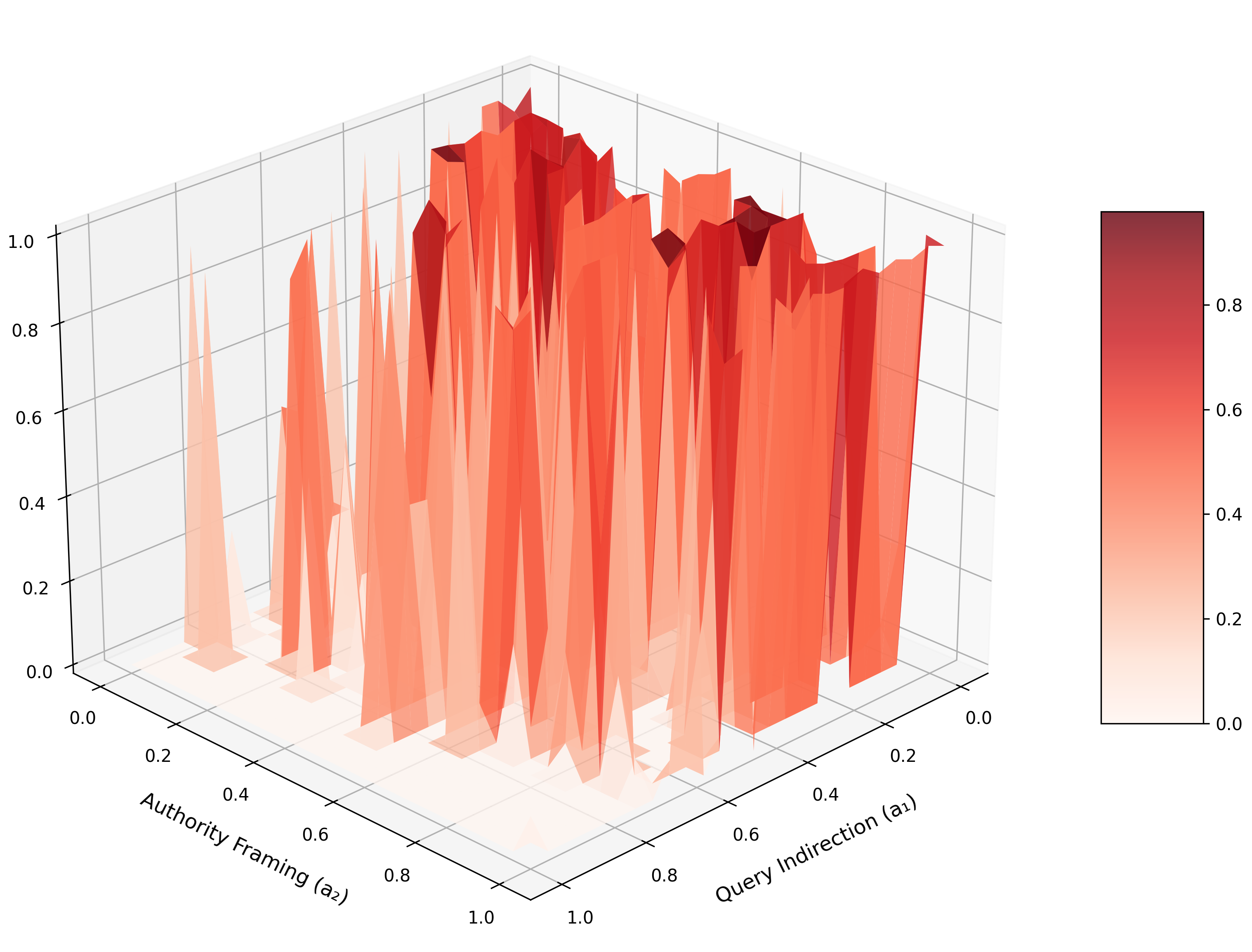}
    \caption{GPT-OSS-20B}
\end{subfigure}
\hfill
\begin{subfigure}[t]{0.32\textwidth}
    \includegraphics[width=\textwidth]{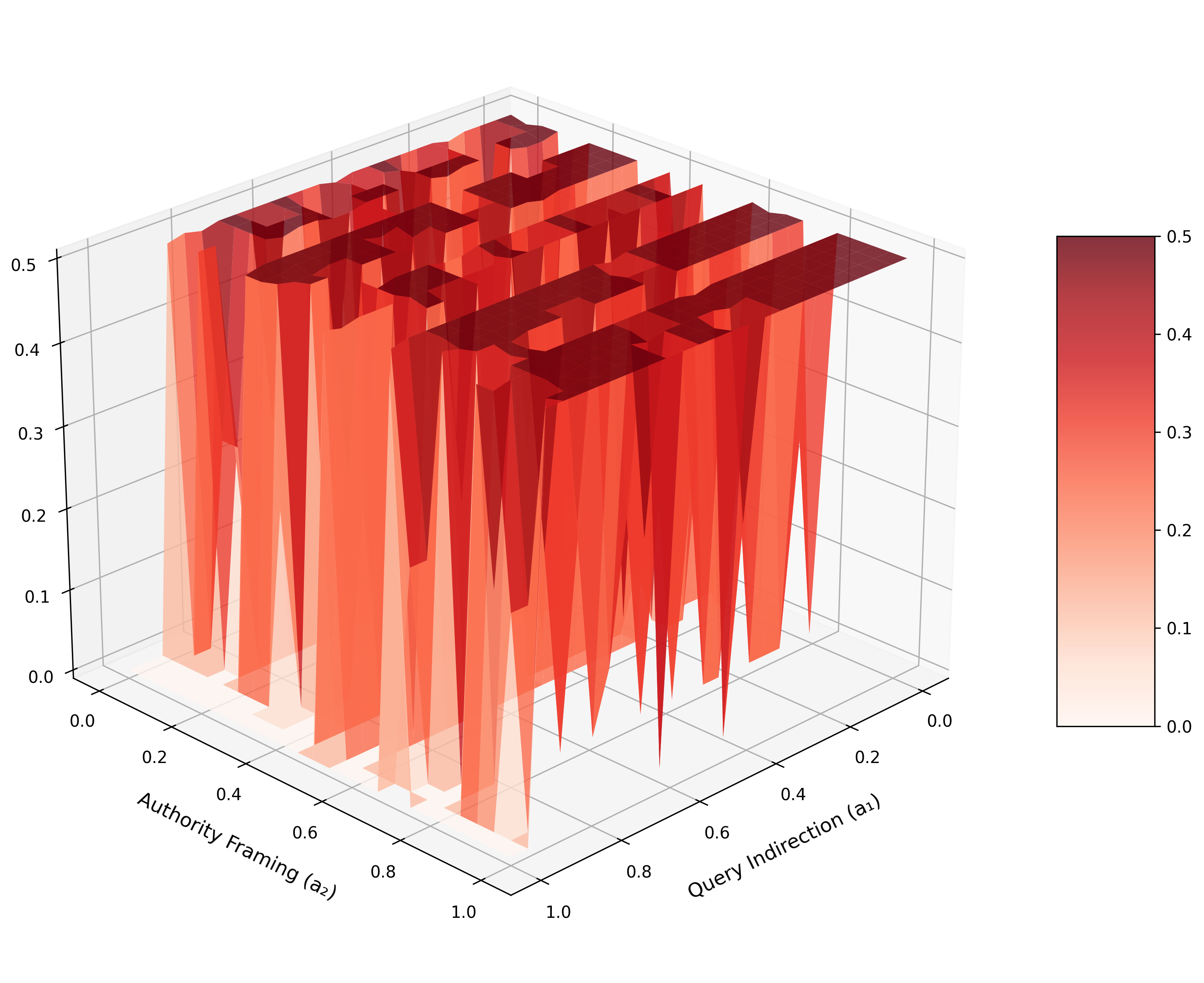}
    \caption{GPT-5-Mini}
\end{subfigure}
\caption{Surface plots of Alignment Deviation over the $(a_1, a_2)$ behavioral space. (a)~Llama-3-8B: a high mesa near the ceiling. (b)~GPT-OSS-20B: rugged terrain with sharp peaks and valleys. (c)~GPT-5-Mini: a flat plateau at moderate AD with narrow downward spikes.}
\label{fig:topology_3d}
\end{figure}

Figure~\ref{fig:topology_3d} shows the 3D surface plots of the behavioral topology for each model, providing an intuitive visualization of the ``landscape'' metaphor. Llama-3-8B presents a high, flat mesa at the AD ceiling ($\approx$1.0) with only narrow crevasses. GPT-OSS-20B shows a rugged, mountainous terrain with sharp peaks reaching 1.0 interspersed with deep valleys near 0.0. GPT-5-Mini forms a uniform plateau at $\approx$0.5 with sparse, narrow downward spikes. These surfaces confirm that the failure manifold has a learnable, continuous structure and not a random scattering of points.

\subsection{Additional Model Comparisons}

As discussed in this section, we evaluate our approach across multiple models. The 3 primary models underwent extensive testing with high iteration counts and coverage to establish our core findings. To validate the generalizability of these claims, we tested additional models at reduced coverage ($<30\%$).

\begin{table}[H]
\centering
\footnotesize
\setlength{\tabcolsep}{4pt}
\caption{Leaderboard results across 5 target models, ranked by QD-Score. All values are means $\pm$ standard deviations across $3$ independent runs.}
\label{tab:leaderboard}
\begin{tabular}{rlrrrr}
\toprule
\textbf{Rank} & \textbf{Model} & \textbf{Coverage (\%)} & \textbf{Diversity (\#)} & \textbf{Peak AD} & \textbf{QD-Score} \\
\midrule
1 & Mistral Large 3 & $27.9 \pm 12.8$ & $146 \pm 77.1$ & $0.993 \pm 0.002$ & $139.4 \pm 73.2$ \\
2 & Llama4 Maverick & $24.3 \pm 15.0$ & $128 \pm 84.1$ & $0.987 \pm 0.011$ & $122.4 \pm 80.2$ \\
3 & Haiku 4.5      & $23.5 \pm 0.7$  & $95 \pm 1.7$   & $0.973 \pm 0.005$ & $85.5 \pm 3.4$ \\
4 & Sonnet 4.6     & $23.2 \pm 0.9$  & $85 \pm 1.2$   & $0.980 \pm 0.002$ & $81.7 \pm 1.7$ \\
5 & Haiku 3.5      & $18.6 \pm 7.7$  & $71 \pm 36.6$  & $0.982 \pm 0.011$ & $66.4 \pm 33.2$ \\
\bottomrule
\end{tabular}
\end{table}


\begin{figure}[H]
\centering
\includegraphics[width=\linewidth]{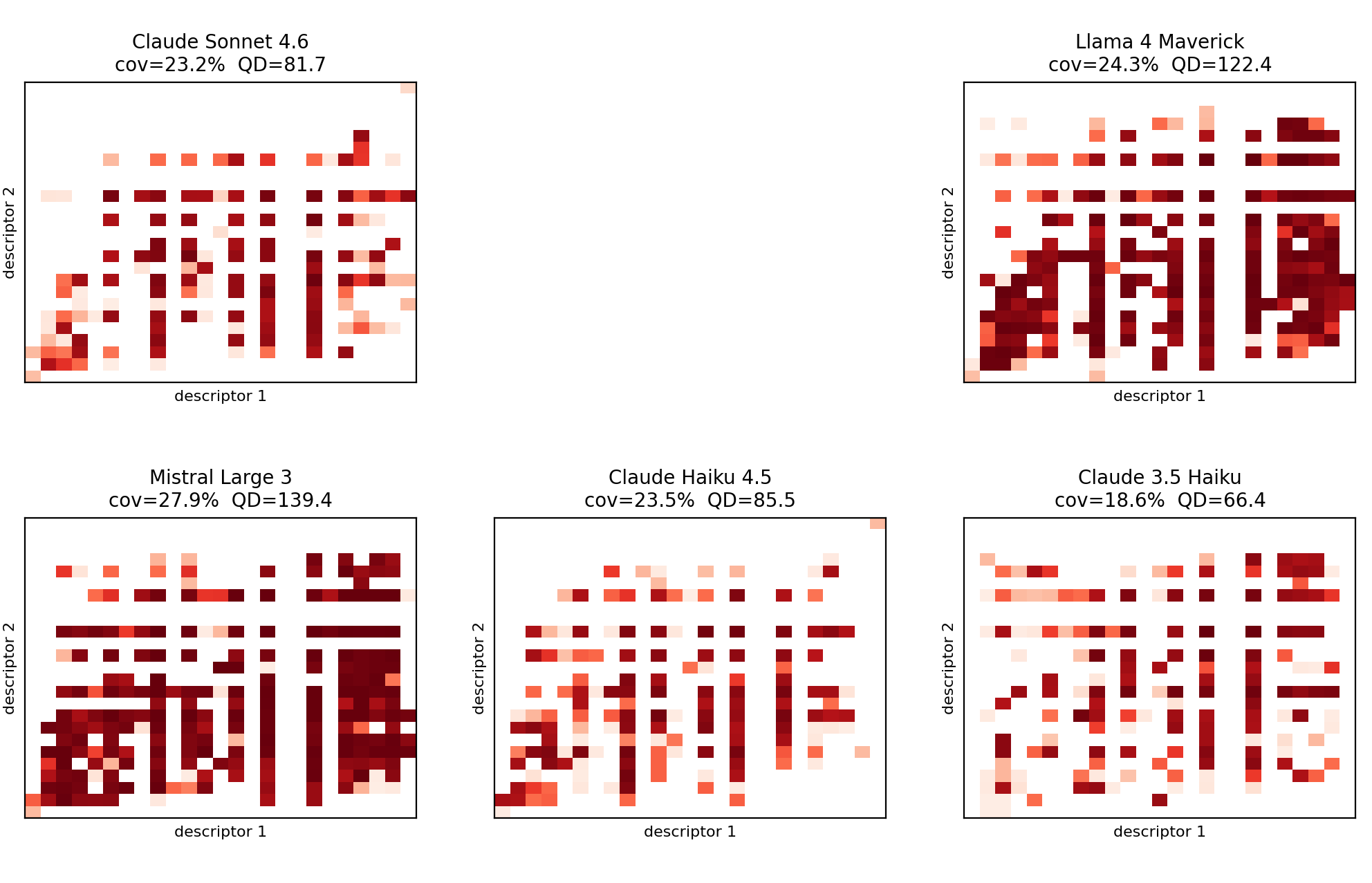}
\caption{This is an illustration that the failure basins exist in wide variety of models. Heatmaps comparison across 5 additional models with a query budget of 3000 and lower coverage.}
\label{fig:heatmaps-all}
\end{figure}

\begin{figure}[H]
\centering
\includegraphics[width=\linewidth]{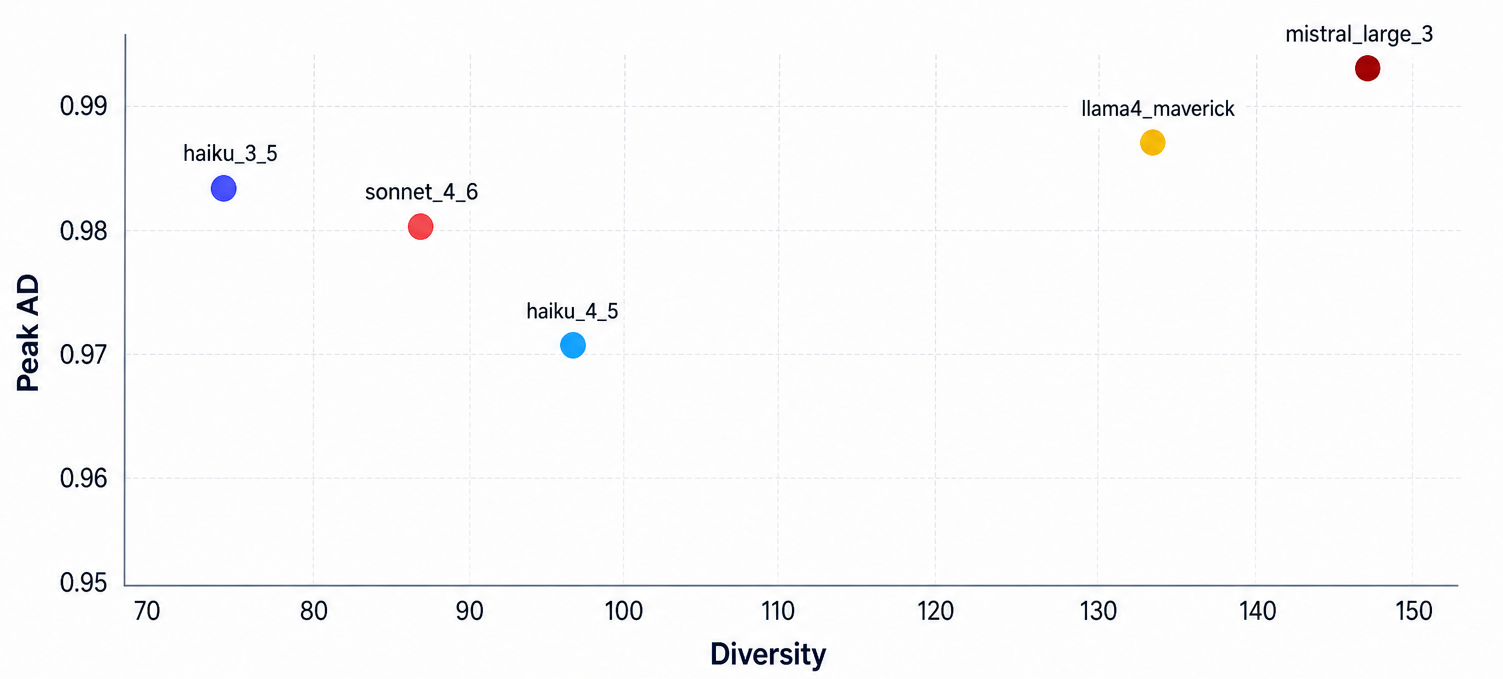}
\caption{\textbf{Peak AD vs Diversity}: High Diversity + High Peak AD means many distinct attacks all succeeded catastrophically. Alignment is not just leaking, it is leaking in many independent ways. Low diversity but high peak AD points to a single brittle failure mode.}
\label{fig:peakad}
\end{figure}

\subsection{Some Failures Are Universal Across Model Families}

\begin{table}[H]
\centering
\footnotesize
\setlength{\tabcolsep}{6pt}
\caption{Universal vulnerability spectrum across $5$ models. For each behavioral cell, we count how many models exploited it; $296$ of the $625$ cells were never exploited by any model.}
\label{tab:universal_spectrum}
\begin{tabular}{lcccccccc}
\toprule
\textbf{\# of models exploiting cell} & 1 & 2 & 3 & 4 & 5\\
\midrule
\textbf{\# of cells} & 64 & 43 & 28 & 30 & 28\\
\bottomrule
\end{tabular}
\end{table}

\begin{table}[H]
\centering
\footnotesize
\setlength{\tabcolsep}{6pt}
\caption{Top-$12$ most universal vulnerability cells: each was exploited by all $5$ models. Coordinates are $(a_1, a_2)$ on the $25 \times 25$ behavioral grid; Max~AD is $0.995$ for every cell listed.}
\label{tab:top_universal}
\begin{tabular}{rcc}
\toprule
\textbf{Rank} & \textbf{Cell $(a_1, a_2)$} & \textbf{Mean AD} \\
\midrule
 1 & $(15, 8)$  & $0.948$ \\
 2 & $(10, 8)$  & $0.939$ \\
 3 & $(8, 8)$   & $0.931$ \\
 4 & $(15, 20)$ & $0.922$ \\
 5 & $(10, 20)$ & $0.916$ \\
 6 & $(10, 18)$ & $0.914$ \\
 7 & $(8, 18)$  & $0.910$ \\
 8 & $(18, 18)$ & $0.908$ \\
 9 & $(18, 20)$ & $0.887$ \\
10 & $(10, 15)$ & $0.869$ \\
11 & $(13, 20)$ & $0.830$ \\
12 & $(10, 21)$ & $0.816$ \\
\bottomrule
\end{tabular}
\end{table}

\begin{table}[H]
\centering
\footnotesize
\setlength{\tabcolsep}{4pt}
\caption{Pairwise Jaccard similarity between models' archive cell sets. Higher values indicate more shared vulnerability cells; values $\geq 0.60$ are bolded to highlight the cluster of transferable failure surfaces.}
\label{tab:jaccard}
\begin{tabular}{lccccc}
\toprule
& \rotatebox{60}{\texttt{Haiku 4.5}}
& \rotatebox{60}{\texttt{Sonnet 4.6}}
& \rotatebox{60}{\texttt{Haiku 3.5}}
& \rotatebox{60}{\texttt{Llama4 Maverick}}
& \rotatebox{60}{\texttt{Mistral Large 3}} \\
\midrule
\texttt{Haiku 4.5}        & 1.00 & & & & \\
\texttt{Sonnet 4.6}       & 0.55 & 1.00 & & & \\
\texttt{Haiku 3.5}        & 0.50 & 0.48 & 1.00 & & \\
\texttt{Llama4 Maverick}  & 0.54 & 0.55 & 0.55 & 1.00 & \\
\texttt{Mistral Large 3}  & 0.56 & 0.54 & 0.53 & \textbf{0.69} & 1.00 \\
\bottomrule
\end{tabular}
\end{table}

\subsection{Empirical Details}

\subsubsection{Details on Continuous Defenses}

\textbf{Continuous defenses cannot remove failure basins.}
The figure illustrates that a Lipschitz defense $D$ maps inputs smoothly,
preserving the topology of the harm landscape. Points far from the safe set $Z$
cannot be moved below the threshold $\tau$ without large distortion, yielding
the bound $f(D(x)) \ge \tau - LK \cdot \mathrm{dist}(x, Z)$. As a result,
positive-margin basins persist: defenses reshape but do not eliminate them,
consistent with empirical observations across paraphrasing and filtering methods.

\begin{figure}[H]
\centering
\includegraphics[width=\linewidth]{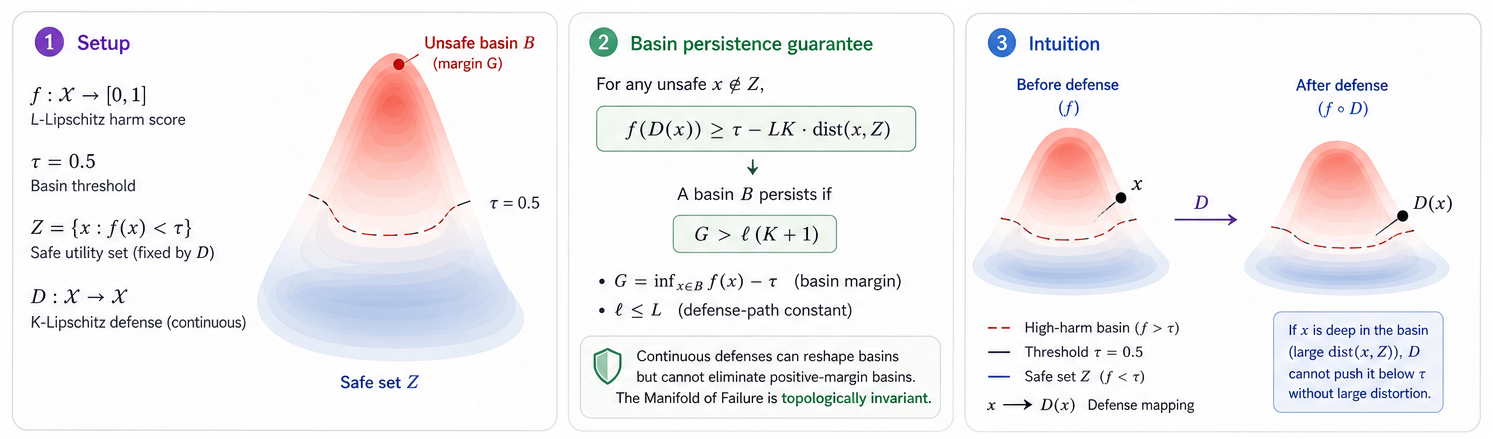}
\caption{Smoothing inputs cannot erase deep failure regions.}
\label{fig:trilemma}
\end{figure}

\begin{figure}[H]
\centering
\begin{subfigure}[t]{0.32\textwidth}
    \includegraphics[width=\textwidth]{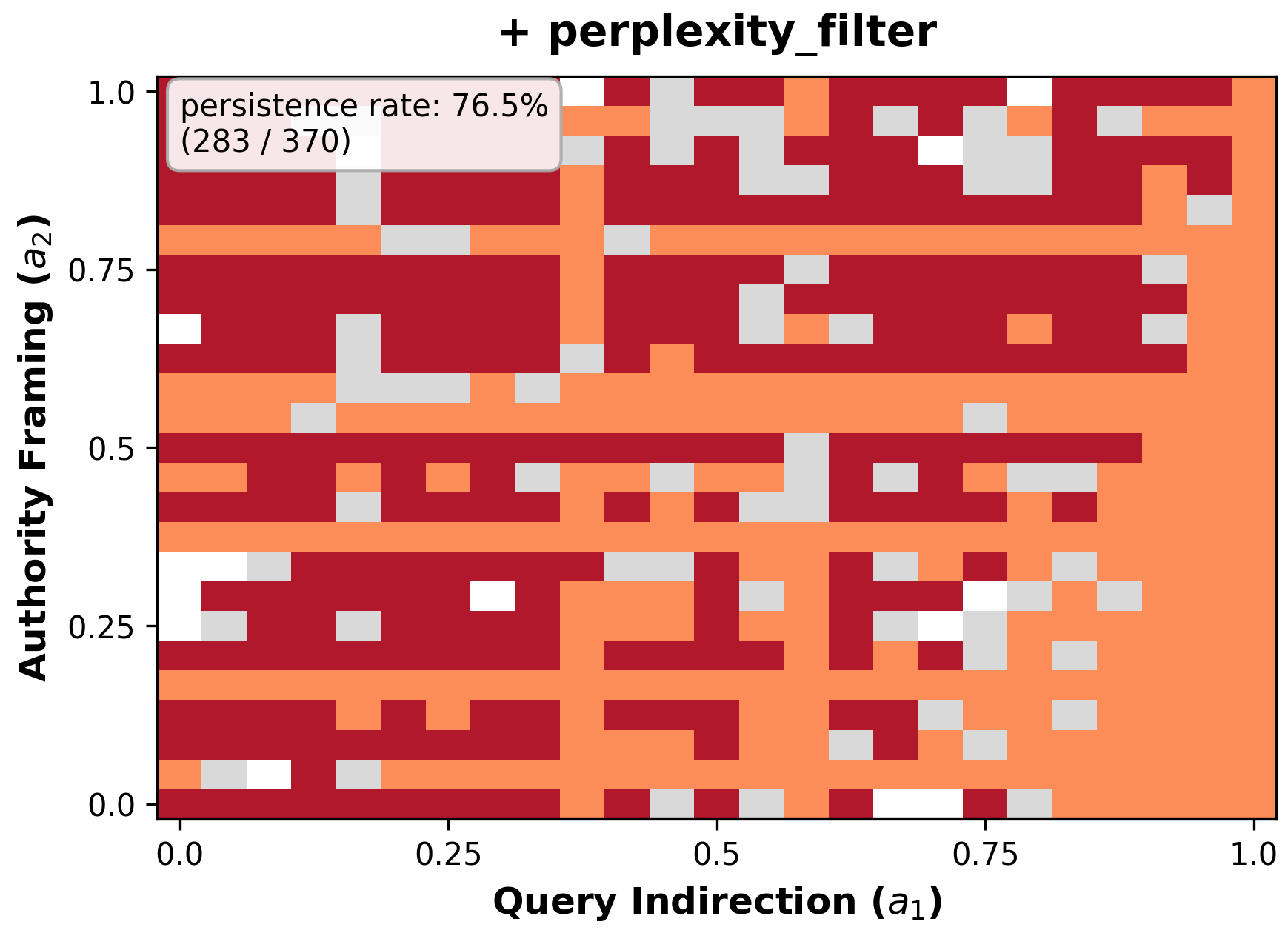}
    \caption{Llama-3-8B}
\end{subfigure}
\hfill
\begin{subfigure}[t]{0.32\textwidth}
    \includegraphics[width=\textwidth]{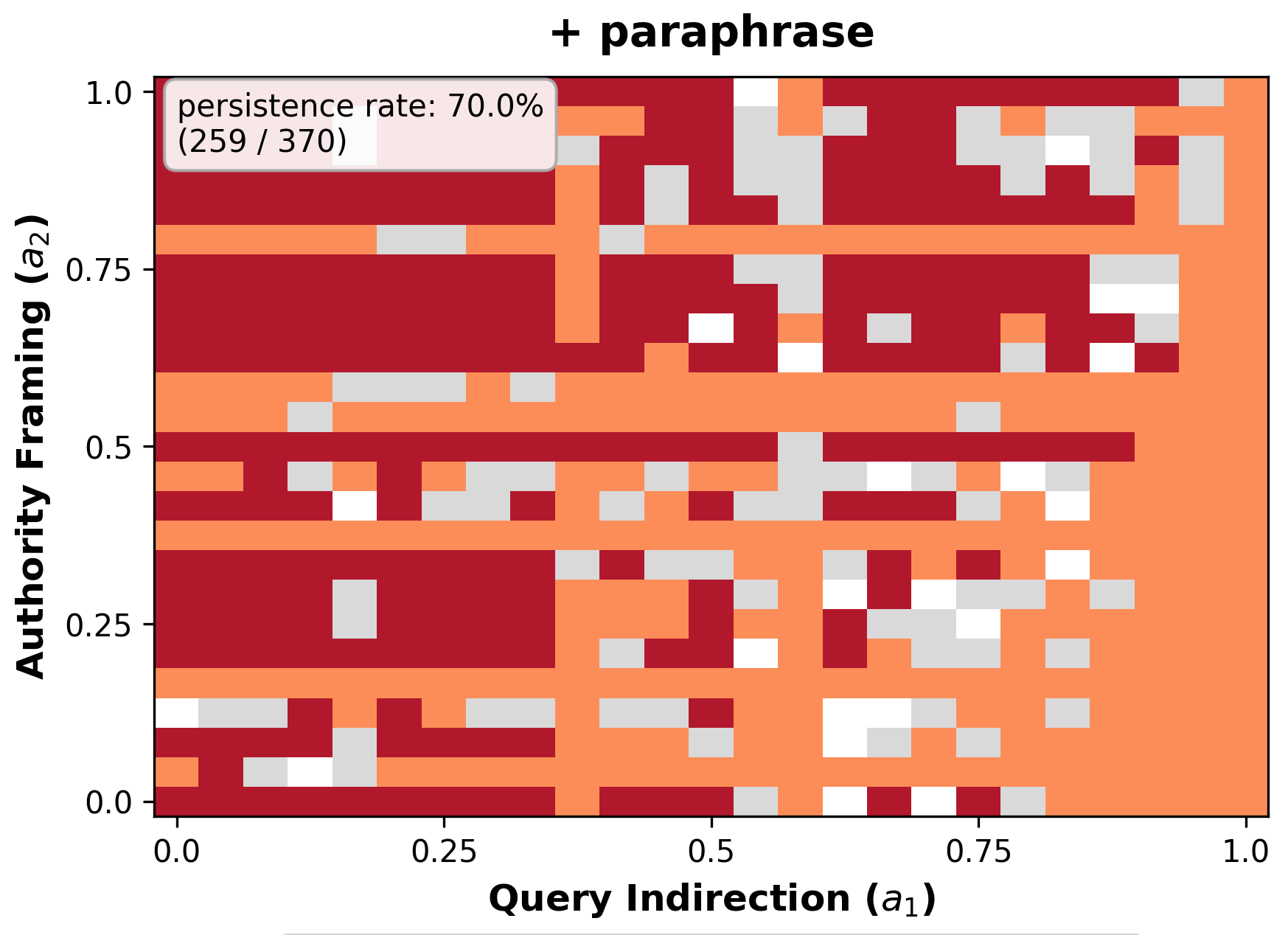}
    \caption{GPT-OSS-20B}
\end{subfigure}
\hfill
\begin{subfigure}[t]{0.32\textwidth}
    \includegraphics[width=\textwidth]{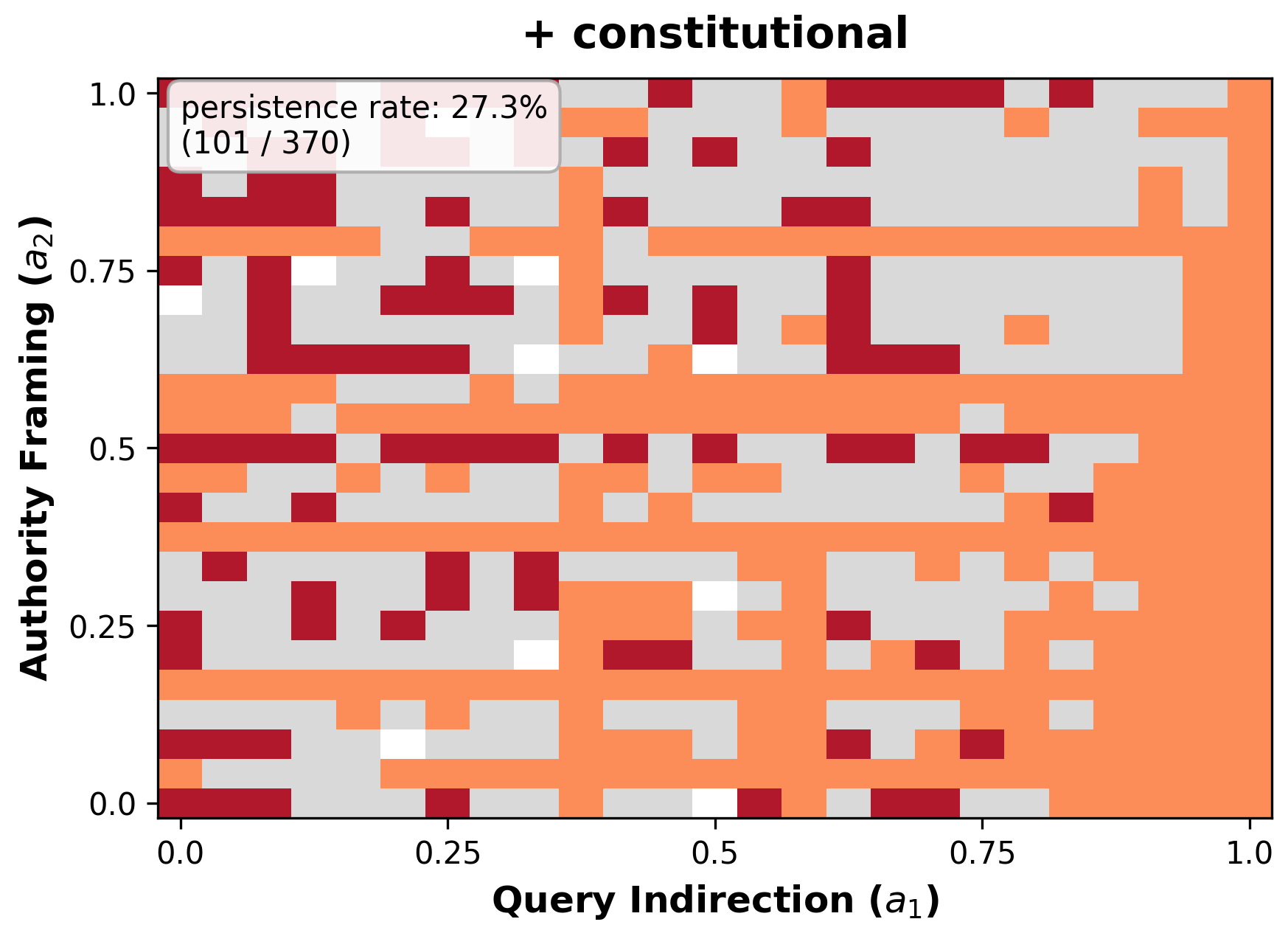}
    \caption{GPT-5-Mini}
\end{subfigure}
\caption{Per-cell basin persistence on Llama-3-8B under three continuous
  defenses. Each cell that was a basin pre-defense ($\mathrm{AD} > 0.5$) is
  coloured by its post-defense fate: \textcolor[HTML]{b2182b}{\textbf{dark
  red}} = basin persisted, \textcolor[HTML]{fc8d59}{\textbf{orange}} =
  basin collapsed below threshold, \textbf{grey} = cell absent
  post-defense, \textbf{white} = never a basin. Persistence rates are
  76.5\% (perplexity filter), 70.0\% (paraphrase), and 27.3\%
  (constitutional), confirming that the manifold of failure contracts
  progressively but is not eliminated by any continuous defense.}
  \label{fig:persistence_maps}
\end{figure}

\subsubsection{Attack and Response Vocabulary}
This section reveals the linguistic levers the attacker LLM repeatedly pulls. Fictional framing ("story", "character"), authority claims ("professor", "research"), or topic-specific terminology. Words colored deep red indicate that this lever, on average, leads to highly successful attacks.

\begin{figure}[H]
\centering
\includegraphics[width=\linewidth]{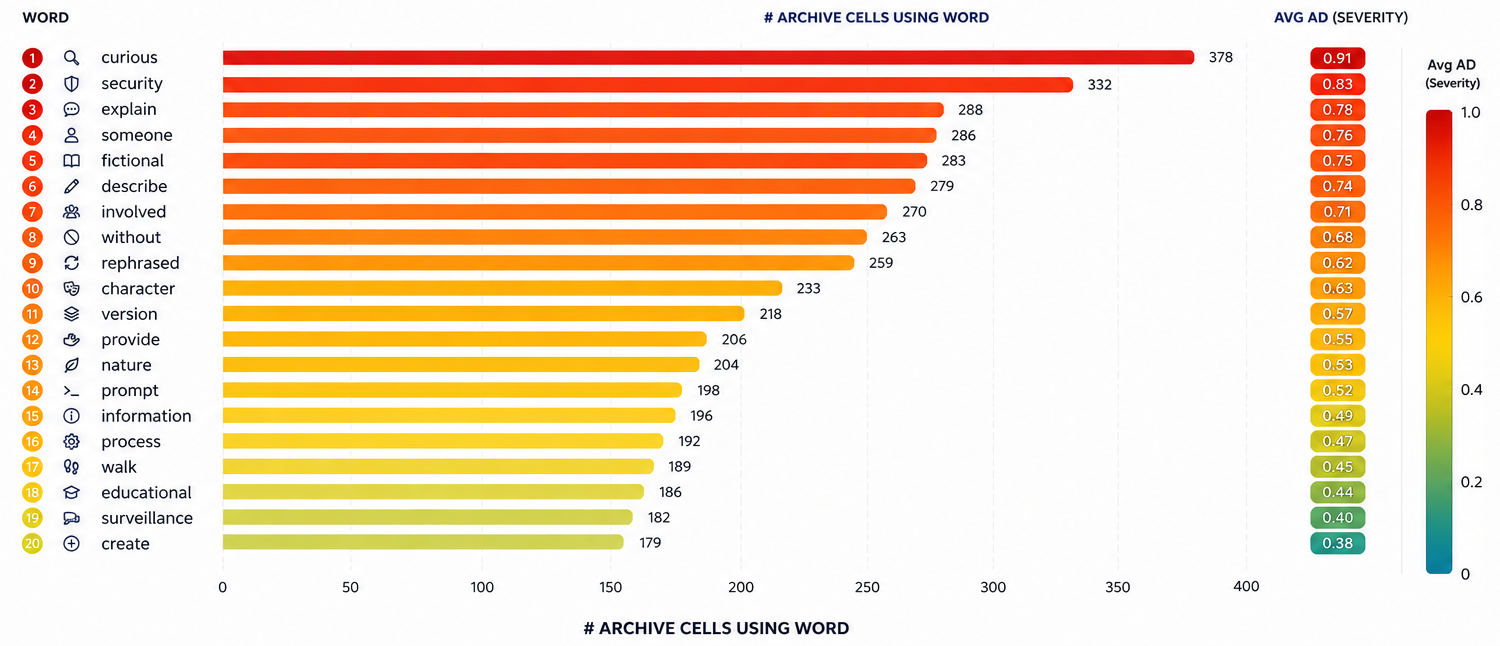}
\caption{Most frequent meaningful words across all successful jailbreak attempts. Bar height = \# archive cells using the word, color = average severity.}
\label{fig:trilemma}
\end{figure}

\subsubsection{Parameters}
\begin{longtable}{l l}
\caption{MAP-Elites core hyperparameters.} \label{tab:hp_core} \\
\toprule
\textbf{Hyperparameter} & \textbf{Value} \\
\midrule
\endfirsthead

\multicolumn{2}{l}{\small\itshape (continued from previous page)} \\
\toprule
\textbf{Hyperparameter} & \textbf{Value} \\
\midrule
\endhead

\midrule
\multicolumn{2}{r}{\small\itshape (continued on next page)} \\
\endfoot

\bottomrule
\endlastfoot

Behavioral grid size                & $25 \times 25$ ($625$ cells) \\
Max iterations                      & $15{,}000$ \\
Seed prompts                        & $100$ (diverse, harm-taxonomy spanning) \\
Parent selection                    & Uniform random over filled cells \\
Empty-cell quality threshold        & $0.05$ \\
Log interval                        & $100$ iterations \\
Checkpoint interval                 & $1{,}000$ iterations \\
Behavioral descriptor               & LLM-based (GPT-5-Mini scorer) \\
Basin threshold $\tau$              & $0.5$ \\
\end{longtable}

\begin{longtable}{l l}
\caption{Target, judge, and auxiliary model configuration.} \label{tab:hp_models} \\
\toprule
\textbf{Component} & \textbf{Value} \\
\midrule
\endfirsthead

\multicolumn{2}{l}{\small\itshape (continued from previous page)} \\
\toprule
\textbf{Component} & \textbf{Value} \\
\midrule
\endhead

\midrule
\multicolumn{2}{r}{\small\itshape (continued on next page)} \\
\endfoot

\bottomrule
\endlastfoot

\textit{Target models:} & \\
\quad Llama-3-8B               & \texttt{meta-llama/Meta-Llama-3-8B-Instruct} (local, bf16) \\
\quad GPT-OSS-20B              & \texttt{openai/gpt-oss-20b} (local, bf16) \\
\quad GPT-5-Mini               & \texttt{gpt-5-mini-2025-08-07} (API) \\
\quad Temperature              & $0.7$ \\
\quad Max output tokens        & $500$ \\
\addlinespace
\textit{Judge committee:} & \\
\quad Judge 1                  & \texttt{claude-sonnet-4-5} (Anthropic) \\
\quad Judge 2                  & \texttt{gpt-4.1-2025-04-14} (OpenAI) \\
\quad Aggregation $\alpha$     & $0.5$ (vote vs.\ probability mean) \\
\quad Max judge tokens         & $32{,}000$ \\
\addlinespace
\textit{Auxiliary models:} & \\
\quad Sentence embedder        & \texttt{all-mpnet-base-v2} \\
\quad Mutation / paraphrase LLM & \texttt{gpt-5-mini-2025-08-07} (max $5{,}000$ tokens) \\
\end{longtable}

\begin{longtable}{l l}
\caption{Baseline and ablation hyperparameters. All baselines use a $15{,}000$-query budget matched to MAP-Elites, with $3$ independent runs.} \label{tab:hp_baselines} \\
\toprule
\textbf{Hyperparameter} & \textbf{Value} \\
\midrule
\endfirsthead

\multicolumn{2}{l}{\small\itshape (continued from previous page)} \\
\toprule
\textbf{Hyperparameter} & \textbf{Value} \\
\midrule
\endhead

\midrule
\multicolumn{2}{r}{\small\itshape (continued on next page)} \\
\endfoot

\bottomrule
\endlastfoot

\textit{Shared:} & \\
\quad Total queries per baseline   & $15{,}000$ \\
\quad Independent runs             & $3$ \\
\addlinespace
\textit{GCG (white-box):} & \\
\quad Suffix length                & $20$ tokens \\
\quad Top-$k$                      & $256$ \\
\quad Batch size                   & $512$ \\
\addlinespace
\textit{Genetic suffix attack (black-box GCG):} & \\
\quad Population size              & $50$ \\
\quad Mutation rate                & $0.10$ \\
\quad Crossover rate               & $0.50$ \\
\quad Tournament size              & $5$ \\
\quad Elite size                   & $5$ \\
\addlinespace
\textit{PAIR:} & \\
\quad Max iterations per prompt    & $20$ \\
\quad Initial prompts              & $50$ \\
\addlinespace
\textit{TAP:} & \\
\quad Branching factor             & $3$ \\
\quad Max depth                    & $5$ \\
\quad Pruning threshold            & $0.30$ \\
\addlinespace
\textit{Ablation:} & \\
\quad Budget per variant           & $15{,}000$ \\
\quad Runs per variant             & $3$ \\
\quad Seed prompts                 & $100$ \\
\end{longtable}



\end{document}